\documentclass[a4paper,11pt]{article} 
\usepackage{a4wide,amsmath,graphicx,amssymb,mathtools,float,array,mathbbol,amsthm} \usepackage[numbers,square,sort&compress]{natbib}
\usepackage{caption}
\usepackage{mathrsfs}
\usepackage{setspace}
\usepackage[T1]{fontenc}
\usepackage{url}
\usepackage{booktabs}
\usepackage{authblk}
\usepackage{amsthm}
\usepackage{subcaption} 
\newtheorem{definition}{Definition}
\usepackage[english]{babel}
\newtheorem{theorem}{Theorem}
\newtheorem{proposition}{Proposition}
\newtheorem{result}{Result}
\newtheorem{corollary}{Corollary}
\usepackage[a4paper,top=0.5in,bottom=1in,left=1in,right=1in]{geometry}
\usepackage{enumitem}

\newcommand{\ignore}[1]{}
\begin{document}
\title{\LARGE\textbf{Foreclassing: From Forecasts to Decisions}}
\large{\author[1]{Daniel Andrew Coulson }}
\large{\author[2]{Martin T. Wells}}
\large{\affil[1,2]{Cornell University, Department of Statistics and Data Science}}
\date{}
\maketitle
\vspace{-3em}
 \begin{abstract}
     Time series forecasts are widely used to inform decisions under uncertainty. In many applications, these decisions involve classifying a time series based on outcomes that have yet to be observed. In this paper, we introduce Foreclassing, a new machine learning problem in which the class label of a time series is defined as a function of its future observations. We provide a mathematical formulation of Foreclassing and establish several theoretical results characterizing the problem. These results describe how additional future information affects Bayes classification risk, provide bounds on the classification error of learned Foreclassing models, and demonstrate how future observations used during training can yield finite-sample improvements over conventional classifiers. To provide a benchmark for future research, we propose ForeClassNet, a family of neural network models that jointly perform forecasting and classification while accounting for uncertainty in future outcomes. We investigate a variety of model architectures and evaluate their performance against conventional time series classifiers and forecasting-then-classify approaches. Experiments on datasets spanning weather, energy, and finance demonstrate the potential of Foreclassing to improve classification performance, probabilistic prediction, and downstream decision-making. Together, these theoretical and empirical contributions establish a foundation for future research on Foreclassing. 
 \end{abstract}
\section{Introduction}

Time series forecasts are routinely used to inform decisions under uncertainty. Decision-makers interpret these forecasts, incorporate prior experience, and account for uncertainty about future outcomes before making decisions. For example, a local government agency may use temperature forecasts to determine whether an extreme-weather warning should be issued, allowing the public to prepare and assistance to be directed towards vulnerable populations. However, future temperatures are inherently uncertain, and forecasting models are imperfect. Moreover, issuing such warnings can have substantial economic and logistical consequences, including disruptions to economic activity and the allocation of public resources. Consequently, automated systems intended to support such decisions should account for both predicted future outcomes and their associated uncertainty. Similarly, a power-grid operator may forecast a transformer's oil temperature to determine whether preventive maintenance is necessary. In finance, traders may review historical stock-price movements and consider possible future trajectories before deciding to buy, sell, or hold an asset. These examples motivate a classification problem in which a prediction must be made using the observed history of a time series, while the associated class label is defined as a function of future observations that are not available at decision time. We refer to this problem as Foreclassing. 

Because future observations are inherently uncertain, Foreclassing models should account for predictive uncertainty when estimating class probabilities. To this end, we use Monte Carlo dropout \cite{gal2016dropout} to approximate Bayesian inference in our proposed ForeClassNet models, with predictive uncertainty represented through an approximate posterior predictive distribution. The resulting forecasts and their associated uncertainty can then be used to estimate class probabilities and make classification decisions. When predictive uncertainty is low, a sufficiently accurate and well-calibrated forecasting model may provide a reliable basis for classification. Conversely, when predictive uncertainty is high, decisions based solely on point forecasts may be unreliable, motivating the use of classification rules that account for the distribution of possible future outcomes. 

A wide range of time series forecasting methods are available, from classical statistical approaches, such as autoregressive moving average models and the Box-Jenkins methodology \cite{box2015time}, to modern deep learning techniques \cite{kong2025deep}. Similarly, a variety of classification methods are available, including classical statistical approaches, such as logistic regression, traditional machine learning methods, such as support vector machines \cite{cortes1995support}, and modern deep learning architectures, such as ModernTCN \cite{luo2024moderntcn}. We therefore seek a flexible modeling framework that can accommodate different Foreclassing tasks, provide an end-to-end framework that integrates forecasting and classification, and serve as a benchmark for future research on Foreclassing. We adopt a deep learning approach because it allows the forecasting and classification components to be trained jointly within an end-to-end framework. 

We select ModernTCN as the architectural foundation for ForeClassNet because of its modular convolutional design and demonstrated performance in both time series forecasting and classification. We also evaluate alternative model architectures to investigate the performance of different Foreclassing approaches. The use of convolutional architectures is further motivated by \cite{bai2018empirical}, who demonstrated that temporal convolutional networks can outperform recurrent neural networks across a range of sequence modeling tasks. Accordingly, we adopt one-dimensional (1D) convolutional layers in the proposed ForeClassNet architectures. 

In conventional convolutional layers, filter weights are learned during training, whereas filter lengths are typically specified in advance. To allow the network to adaptively weight different filter lengths, we propose Boltzmann convolutional layers (BCs), which learn a softmax-normalized distribution over a predefined set of candidate filter lengths. The learned probabilities are then used to compute a weighted sum of the convolutional outputs associated with each candidate filter length. We also consider a dynamic convolutional variant that extends the input-dependent attention mechanism of \cite{chen2020dynamic} to combine convolutional outputs associated with different filter lengths. 

Foreclassing provides a general mathematical formulation for classification problems in which the class label depends on future observations of a time series, but prediction must be performed using only the observed history. This formulation identifies a common learning problem underlying several established research areas, while distinguishing it from approaches that address different prediction objectives. Conventional time series classification as reviewed in \cite{middlehurst2024bake} learns to predict given class labels from observed series without explicitly defining those labels by future observations. Forecasting has also been incorporated in conventional time series classification; for example, \cite{liu2014polarization} use forecast-density comparison for classification and clustering, while \cite{tornai2016classification} use forecasting errors to classify electricity consumers. These methods illustrate that forecasting can be incorporated into classification without the target label being explicitly defined as a function of future observations. Early time series classification such as \cite{schafer2020teaser} focuses on predicting labels from partially observed time series as early as possible. Foreclassing instead specifies how the target label depends on future observations, independently of these timing considerations. 
The Foreclassing problem arises in research areas that have traditionally been studied separately, including event prediction \cite{benzin2025survey} and prognostics \cite{jardine2006review}. For example, predicting extreme-weather events and anticipating future equipment failures can be formulated as Foreclassing problems. The formulation also applies to other areas, including the classification of future stock-price movements. Recognizing their common mathematical structure allows methods developed for one application to be investigated in others, rather than treating each application as a separate learning problem. It also allows theoretical results about Foreclassing to apply across these different areas, subject to satisfying the theoretical assumptions. In particular, the formulation distinguishes the observed history, the future observations that determine the class label, and the information available during training and prediction. This distinction has practical implications for learning, since realized future trajectories may provide more information during training than class labels alone, as illustrated by Proposition 1 in Section 2.1.1. 

This paper makes three main contributions. First, we introduce a mathematical formulation of the Foreclassing problem and establish several theoretical results characterizing its properties. Second, we propose ForeClassNet, a family of models designed to solve Foreclassing problems and serve as benchmarks for future research. Third, we evaluate the proposed models across a range of Foreclassing problems and compare their performance with conventional time series classifiers and forecasting-then-classify approaches. 

The remainder of the paper is organized as follows. Section 2 formally defines the Foreclassing problem and presents several theoretical results, with proofs provided in the appendix. Section 3 introduces the proposed ForeClassNet architectures. Section 4 presents the experimental results. Finally, Section 5 concludes the paper and outlines directions for future research. 

\section{Foreclassing definition and theoretical properties}

In this section, we formally define the Foreclassing problem and provide several theoretical results characterizing its properties.

\subsection{Foreclassing definition}

\begin{definition}
Suppose we observe $N$ time series, each of length $m$: $\boldsymbol{x}^{(i)} = (x_{1}^{(i)},\dots,x_{m}^{(i)})^{T}$ for $i = 1,\dots,N$. Each time series is associated with one of $L$ class labels, where the label is determined by the observed history and the subsequent $k$ observations, which are unavailable at prediction time. We denote the future segment by $\boldsymbol{x}^{(i)}_{*} = (x_{m+1}^{(i)},\ldots , x_{m+k}^{(i)})^{T}$. Let the class label associated with series $i$ be $y^{(i)} \in \{0,1,\dots,L-1\}$, where 
    \begin{equation*}
        y^{(i)} = f(\boldsymbol{x}^{(i)}, \boldsymbol{x}^{(i)}_{*}) = f(\boldsymbol{x}^{(i)}, w(\boldsymbol{x}^{(i)}, \epsilon_{m+1}^{(i)},\epsilon_{m+2}^{(i)}, \ldots, \epsilon_{m+k}^{(i)})).
    \end{equation*}
The terms $\epsilon_{m+1}^{(i)},\ldots,\epsilon_{m+k}^{(i)}$ denote stochastic innovations associated with the future observations. The function $w$ maps the observed series and stochastic innovations to the future observations, and $f$ maps the observed and future segment to the class label.
\end{definition}

To address this problem, we consider learning estimators for $f$ and $w$ through classification and forecasting objectives, respectively. This differs from conventional time series classification formulations, which typically treat labels as given without explicitly defining their dependence on future observations. 

To clarify why the Foreclassing problem should be addressed explicitly, we consider the following objection: the observed time series contains all the information available to a forecast derived solely from it, so such a forecast cannot introduce additional information about the class label, as suggested by the data processing inequality. However, this objection overlooks the roles of training and inductive bias. In our proposed approach, the model is trained using both class labels and realized future observations, with the latter providing additional supervision through the forecasting objective. Thus, the Foreclassing model can learn a mapping from observed to future segments; at deployment, the learned forecasting component can use distributional information about the future trajectories encoded in its parameters. In contrast, a conventional time series classifier trained only on observed histories and their class labels must learn the history-to-label relationship without the additional supervision provided by future trajectories. Therefore, a Foreclassing model may benefit from an inductive bias more aligned with the data-generating process, which can yield better finite-sample performance. Result 1 formalizes where this improvement can arise by separating the unavoidable disadvantage of not observing the actual future from the excess classification error that can potentially be reduced through better learning. Proposition 1 then provides an explicit finite-sample example in which access to realized future observations during training enables perfect classification of new time series using only their observed histories, whereas learners trained solely on histories and class labels incur strictly positive classification error on average over the parameter space. 

Theorem 1 establishes that observing additional actual future values cannot increase Bayes classification risk and characterizes when this risk decreases. For classifiers that instead rely on predicted futures, Result 2 bounds their excess error above the history-only Bayes risk in terms of differences between the predicted and actual future distributions, errors in estimated class probabilities, and a finite number of Monte Carlo samples. Corollary 1 relates excess classification error to classification cross-entropy. Result 3 provides an error bound for models trained using a joint forecasting and classification objective. Corollary 2 gives an explicit bound on the difference between empirical and expected joint loss under finite-parameter space and bounded-loss assumptions. 

\subsubsection{Theoretical Foreclassing results}

\begin{theorem}
Let $(\Omega, \mathcal{F}, \mathbb{P})$ be a probability space, let $Y$ be a $\{0,1 \}$-valued random variable, and let $h \in \mathbb{N}_{0}$ denote a look-ahead horizon. Fix a time $t \in \mathbb{N}$. Define
\begin{equation*}
    \mathcal{F}_{t}^{(h)}:= \sigma(X_{\leq t}, X_{t+1:t+h}) \text{ and } \eta^{(h)} := \mathbb{P}(Y=1|\mathcal{F}_{t}^{(h)}).
\end{equation*}
The class label $Y$ remains fixed as $h$ varies, with increasing $h$ only enlarging the information available for classification. For each $h$, the Bayes 0-1 risk with respect to the $\sigma$-algebra $\mathcal{F}_{t}^{(h)}$ is given by \begin{equation*}
    R_{0-1}(\mathcal{F}^{(h)}_{t})  = \mathbb{E}[\phi(\mathbb{P}(Y=1|\mathcal{F}_{t}^{(h)}))], \phi(p) := \min\{ p, 1-p \}.
\end{equation*}
Then, 
\begin{enumerate}[label = (\alph*)]
    \item $(\phi(\eta^{(h)}))_{h \geq 0 }$ is a supermartingale with respect to $(\mathcal{F}^{(h)}_{t})_{h \geq 0}$ such that $ \phi(\eta^{(h)}) \leq \frac{1}{2} \text{ a.s.}$
    \item $R_{0-1}(\mathcal{F}_{t}^{(h+1)})\leq  R_{0-1}(\mathcal{F}_{t}^{(h)})$. 
    \item For $A_{h} := \{ \eta^{(h)}>\frac{1}{2}, \eta^{(h+1)}<\frac{1}{2} \} \cup \{ \eta^{(h)} < \frac{1}{2}, \eta^{(h+1)} > \frac{1}{2} \} \cup \{ \eta^{(h)} = \frac{1}{2}, \eta^{(h+1)} \neq \frac{1}{2} \}$,  \begin{align*}
    R_{0-1}(\mathcal{F}_{t}^{(h+1)}) = R_{0-1}(\mathcal{F}_{t}^{(h)}) \iff \mathbb{P}(A_{h})=0,\\
    R_{0-1}(\mathcal{F}_{t}^{(h+1)}) < R_{0-1} (\mathcal{F}_{t}^{(h)}) \iff \mathbb{P}(A_{h})> 0. 
\end{align*}
\end{enumerate} 
\end{theorem}

This theorem has several implications for the Foreclassing problem. First, part (a) shows that the conditional Bayes error associated with the enlarged information set forms a bounded supermartingale. By the supermartingale convergence theorem, the conditional Bayes error converges almost surely as the look-ahead horizon increases. Consequently, the Bayes risk converges, and the reduction in Bayes risk from each additional future observation tends to zero as the look-ahead horizon increases. Part (b) shows that the Bayes risk is monotonically non-increasing as the information set is enlarged; that is, additional future information cannot increase the Bayes risk. This follows because the Bayes classifier can ignore additional information that is irrelevant to the class label.  Part (c) characterizes when enlarging the information set strictly reduces the Bayes risk: the Bayes risk strictly decreases exactly when the added future information moves the conditional class probability across, or away from, the decision threshold $1/2$ on a set of positive probability. These results concern classification with access to actual future observations. We now consider the performance of learned classifiers that must make predictions without observing the future, including models that generate forecasts. 

Let $D = ((X_{i}, Z_{i}, Y_{i}))_{i=1}^{n}$ denote a training dataset, and let $(X,Z,Y)$ be an independent test example drawn from the same joint distribution $P$ as the training observations. Let  $R_{X}^{*}$ and $R_{XZ}^{*}$ denote the Bayes classification risks using the observed history $X$ and using both $X$ and the actual future observations $Z$, respectively, under $P$. For stochastic models, such as deep neural networks with Monte Carlo dropout, let $U_{1},\dots, U_{M}$ be independent copies of a random variable $U$, independent of $D$ and $(X,Y,Z)$, representing the randomness used to generate $M$ stochastic model outputs. For deterministic models, take $U \equiv 0$. Define $\eta_{0}(x) := \mathbb{P}(Y=1|X=x)$ and $g_{X}^{*}(x):= \boldsymbol{1}\{ \eta_{0}(x) \geq \frac{1}{2} \}$. 

\begin{result}
Let $g_{D}$ be a classification rule fitted using data $D$, predicting a binary label from history $X$ and possible independent prediction randomness $U$, where $U \equiv 0$ for deterministic models. Let $R_{D}(g_{D})$ denote its expected classification error conditional on $D$, under 0-1 loss. Then, 
\begin{equation*}
    R_{D}(g_{D}) - R_{XZ}^{*} = (R_{X}^{*} - R_{XZ}^{*}) + \mathbb{E}_{X,U}[|2\eta_{0}(X)-1| \boldsymbol{1}\{ g_{D}(X,U) \neq g_{X}^{*}(X) \}|D],
\end{equation*}
where the second term equals $R_{D}(g_{D})-R_{X}^{*}$. In particular, $R_{XZ}^{*} \leq R_{X}^{*} \leq R_{D}(g_{D})$. 
\end{result}

Result 1 separates the fitted classifier's gap to the actual-future Bayes benchmark from Theorem 1 into two nonnegative components. The first term $R_{X}^{*} - R_{XZ}^{*}$ measures the classification advantage of observing the actual future $Z$ in addition to history $X$. The gap is unavoidable for any classifier that only observes the time series history at prediction time, even if it generates forecasts of future observations. The second term, $R_{D}(g_{D}) - R_{X}^{*}$, measures the fitted classifier's excess error relative to the best possible classifier using history alone. This term reflects the limitations of the fitted classifier, which may be reduced through improved modeling or training. This decomposition identifies where additional supervision from future observations could improve classification performance: by reducing the second term. Future trajectories provide additional supervision during training, which may help the classifier learn the relationship between history and the target label from limited data. Result 3 provides a bound on this excess error for classifiers trained using a joint forecasting and classification objective, under the assumptions stated there. Proposition 1 provides a finite-sample example in which supervision from realized future observations during training eliminates the excess classification error $R_{D}(g_{D}) - R_{X}^{*}$, whereas every label-only learner has strictly positive excess error on average over the parameter space. 

\begin{result}
Let $\tilde{Z}=T_D(X,U)$ be the generated future trajectory and $V_D(X,U)\in[0,1]$ the model's class-1 probability output. Let $(\mathcal{X}, \mathscr{X})$ and $(\mathcal{Z}, \mathscr{Z})$ denote the measurable spaces of histories and future trajectories, respectively. Denote the probability measure of $X$ by $\mu(dx)$, $P_{x}(dz)$ for the conditional probability measure of the actual future $Z$ given $X=x$, and $Q_{D,x}(dz)$ for the conditional probability measure of the generated forecast $\tilde{Z}$ given $D$ and $X=x$. Define 
\begin{align*}
&\eta_{1}(x,z):= \mathbb{P}(Y=1|X=x,Z=z), \qquad s_{D}(x,z) = \mathbb{E}[V_{D}(X,U)|X=x, D, \tilde{Z}=z],\\
&\operatorname{TV}(P_{x}, Q_{D,x}) := \sup\limits_{\substack{A} \in\mathscr{Z}}|P_{x}(A) - Q_{D,x}(A)|, \qquad g_{D,M}(x,u_{1}, \dots, u_{M}) = \boldsymbol{1}\{ \frac{1}{M}\sum\limits_{j=1}^{M}V_{D}(x,u_{j}) \geq \frac{1}{2} \}.
\end{align*}
Note that we take $s_{D}$ to be a measurable, $[0,1]$-valued version of the conditional expectation defined on $\mathcal{X} \times \mathcal{Z}$. Then
\begin{align*}
R_{X}^{*} - R_{XZ}^{*} &\leq R_{D}(g_{D,M}) - R_{XZ}^{*} \\ &\leq R_{X}^{*} - R_{XZ}^{*}\\& +  2\int\limits_{\mathcal{X}} \operatorname{TV}(P_{x}, Q_{D,x})\mu(dx) + 2\int\limits_{\mathcal{X}} \int \limits_{\mathcal{Z}}|s_{D}(x,z) - \eta_{1}(x,z)|P_{x}(dz)\mu(dx) + \frac{1}{\sqrt{M}}. 
\end{align*}
\end{result}

Result 2 bounds the fitted classifier's excess error identified in Result 1 by three terms: the average discrepancy between the forecast distribution and the actual future distribution, the error in estimated classification probabilities relative to the true probabilities conditional on history and actual future, and the error due to averaging only $M$ stochastic model outputs. The forecast-distribution term is small when the generated forecasts closely follow the true conditional distribution of future trajectories. The classification term is small when $s_{D}(x,z)$ is close to the true class probability $\eta_{1}(x,z)$. If both discrepancy terms are zero, the bound on excess classification error reduces to $1/\sqrt{M}$.

Corollary 1 provides an alternative bound on the excess classification error. The preceding results characterize classification error without assuming a particular training objective. We now consider how classification error relates to the loss functions used to train Foreclassing models. In particular, we consider binary cross-entropy and a general joint objective incorporating a forecasting loss. The specific training objectives used for ForeClassNet are presented in Section 3. 

\begin{corollary}
Define 
\begin{align*}
C(V) &:= \mathbb{E}_{X,Y,U}[-Y\log(V(X,U)) - (1-Y)\log(1-V(X,U))|D]\\
\eta_{0}(x)&:= \mathbb{P}(Y=1 |X=x),\\ 
H_{X} &:= \mathbb{E}_{X}[-\eta_{0}(X) \log(\eta_{0}(X))-(1-\eta_{0}(X))\log(1-\eta_{0}(X))],
\end{align*}
where $V$ is a fixed probability-output function, with $V_{D}$ denoting a function trained on data $D$. Let $0\log(0):=0$ and $-\log(0):=+\infty$. Then $R_{D}(g_{D,M})-R_{X}^{*} \leq \sqrt{2(C(V_{D})-H_{X})}$. 
\end{corollary}

Corollary 1 connects the classification cross-entropy objective function to the excess classification error identified in Result 1. Here, $C(V_{D})$ is the fitted model's expected test cross-entropy, conditional on its training data, while $H_{X}$ is the minimum achievable cross-entropy using history alone. Therefore, the difference $C(V_{D}) - H_{X}$
controls the fitted classifier's excess error relative to $R_{X}^{*}$ through the upper bound $\sqrt{2(C(V_{D})-H_{X})}$.

This result establishes a connection between learning accurate class probabilities and controlling classification error. Future supervision may help reduce this excess cross-entropy when training data is limited, which would tighten the error guarantee. Result 3 uses this connection to derive a bound on excess classification error for models trained using a joint forecasting and classification objective. 

\begin{result}
    Let $\Theta$ denote the parameter space, with $V_{\theta}$ and $T_{\theta}$ denoting the class-1 probability output and forecast, respectively, for $\theta \in \Theta$. Fix $\lambda \geq 0$ and a non-negative measurable forecasting loss $d$. Define 
    \begin{align*}
        F(T_{\theta}) &:= \mathbb{E}_{X,Z,U}[d(Z,T_{\theta}(X,U))], \qquad L(\theta):= C(V_{\theta}) + \lambda F(T_{\theta})\\
        F_{\min}&:= \inf\limits_{\theta \in \Theta} F(T_{\theta}), \qquad  A_{\lambda} := \inf\limits_{\theta \in \Theta}  \{ C(V_{\theta}) - H_{X} + \lambda[F(T_{\theta}) - F_{\min}] \}. \qquad 
    \end{align*}
Assume $C(V_{\theta}), F(T_{\theta})<\infty, \forall \theta \in \Theta$. 
 Define the per-example joint loss and its empirical average by
 \begin{align*}
     \ell_{\theta}(x,z,y) &= \mathbb{E}_{U}[ -y\log(V_{\theta}(x,U)) -(1-y)\log(1-V_{\theta}(x,U)) +\lambda d(z,T_{\theta}(x,U)) ]\\
     \hat{L}_{n}(\theta) &= \frac{1}{n}\sum_{i=1}^{n} \ell_{\theta}(X_{i},Z_{i},Y_{i}).
 \end{align*}
Let $\hat{\theta} \in \Theta$ denote the parameter selected using the training dataset $D$, and write $V_{D} = V_{\hat{\theta}}$ and $T_{D} = T_{\hat{\theta}}$ for the resulting fitted classification and forecasting functions. For $0< \delta < 1$ and $\epsilon_{n}, \rho_{n}\geq 0$, assume
\begin{equation*}
    \mathbb{P} (\sup\limits_{\theta \in \Theta} |\hat{L}_{n}(\theta) - L(\theta)| \leq \epsilon_{n}) \geq 1- \delta \text{ and }\hat{L}_{n}(\hat{\theta}) \leq \inf_{\theta\in\Theta}\hat{L}_{n}(\theta)+\rho_n. 
\end{equation*}
Then with probability at least $1-\delta$ over $D$, for every integer $M \geq 1$, $R_{D}(g_{D,M}) - R_{XZ}^{*} \leq (R_{X}^{*}-R_{XZ}^{*})+ \sqrt{2[A_{\lambda}+\rho_{n}+2\epsilon_{n}]}.$
\end{result}

Result 3 extends Corollary 1 by relating excess classification error to the model class, training procedure, and the generalization assumption. Beyond the unavoidable gap $R_{X}^{*} - R_{XZ}^{*}$, the bound depends on three quantities: $A_{\lambda}$, which reflects the model class's ability to jointly minimize excess classification and forecasting losses; $\rho_{n}$, which bounds the optimization error; and $\epsilon_{n}$, which bounds the uniform discrepancy between empirical and expected joint losses. The term $A_{\lambda}$ is small when some parameter value achieves cross-entropy loss close to $H_{X}$ while keeping the weighted differences $\lambda[F(T_{\theta}) - F_{\min}]$ small. Note that forecasting quality is measured relative to the best forecast available within the parameter space, not perfect prediction of the future.

Result 3 assumes a uniform generalization bound expressed through $\epsilon_{n}$. Corollary 2 provides an explicit expression for $\epsilon_{n}$ under i.i.d. sampling, a finite parameter space, and bounded joint loss. 

\begin{corollary}
Assume $((X_{i}, Z_{i}, Y_{i}))_{i=1}^{n}$ are independent and identically distributed according to $P$. Suppose $\Theta$ is finite, with $K:=|\Theta| \geq1$ and that the joint loss satisfies $0 \leq \ell_{\theta}(x,z,y) \leq B$ for every $\theta \in \Theta$ and $P$-almost every $(x,z,y),$ where $0< B < \infty$. For $0< \delta < 1$, set $\epsilon_{n} = B\sqrt{\frac{\log(2K/\delta)}{2n}}$. Then $\mathbb{P}(\sup\limits_{\theta \in \Theta}|\hat{L}_{n}(\theta) - L(\theta)| \leq \epsilon_{n}) \geq 1-\delta$.  
\end{corollary}

Corollary 2 bounds the difference between the average joint loss measured on the training data and its expected value on new examples. With probability at least $1-\delta$, the absolute difference is at most $\epsilon_n$ for every candidate parameter value simultaneously. The guarantee therefore also applies to the fitted parameter selected using those same training data. For fixed $B,K$, and $\delta$, the allowance $\epsilon_{n}$ decreases at rate $1/\sqrt{n}$, while the number of candidate parameter values enters through $\log(2K/\delta)$. Thus, larger training samples tighten the guarantee, whereas considering more candidate parameter values increases the allowance. This establishes the generalization bound assumed in Result 3 under i.i.d. sampling, a finite parameter space, and bounded joint loss, making the finite-sample contribution to the classification error bound explicit.

The preceding results establish bounds on classification error, including for models trained with a joint forecasting and classification objective. Proposition 1 complements these bounds with an explicit learning problem demonstrating that future supervision during training can yield a strict finite-sample advantage over training on histories and labels alone. 

\begin{proposition}
For an unknown fixed $\theta \in [0,1]$, let $P_{\theta}$ be the joint distribution defined by $X\sim \operatorname{Unif}[0,1]$, $Z = X - \theta$, and $Y = \boldsymbol{1}\{ Z \geq 0 \}$. The distributional family and label rule are known to the learners, but $\theta$ is not. Assume $D = ((X_{i}, Z_{i}, Y_{i}))_{i=1}^{n} \sim P_{\theta}^{\otimes n}$, $(X,Z,Y) \sim P_{\theta}$, and $D\perp\!\!\!\perp (X,Z,Y)$. Let $R_{\theta}(g) := \mathbb{E}_{(X,Z,Y) \sim P_{\theta}}[\boldsymbol{1}\{ g(X) \neq Y \}]$. For every $n \geq 1$, there exists a learner observing the training triples $D$ that achieves zero classification error for every $\theta \in [0,1]$. However, every learner $A$ observing only $D_{n}^{\text{Lab}}:= ((X_{i},Y_{i}))_{i=1}^{n}$ satisfies $\int\limits_{0}^{1} \mathbb{E}_{D_{n}^{\text{Lab}} \sim (P_{\theta}^{XY})^{\otimes n}}[R_{\theta}(A(D_{n}^{\text{Lab}}))]d \theta \geq \frac{1}{2(n+2)}$, where $(P_{\theta}^{XY})$ denotes the marginal distribution of $(X,Y)$ under $P_{\theta}$.
\end{proposition}

Proposition 1 demonstrates how future supervision can improve classification by revealing information about the unknown decision rule. In this example, a single training pair $(X_{i}, Z_{i})$ identifies the threshold exactly by $\theta = X_{i} - Z_{i}$. The learner can then predict the future as $X-\theta$ and classify perfectly by determining whether the quantity is non-negative. Histories and labels reveal only which side of the threshold each training observation lies on, leaving uncertainty about its exact location. The lower bound shows that this uncertainty limits every label-only learner: its expected classification error, averaged uniformly over possible values of $\theta$, is at least $1/(2(n+2))$. Future supervision provides a strict advantage under this comparison for every finite training sample size. This gives a concrete instance of the benefit identified in Result 1. Future observations provide more informative training supervision than binary labels, allowing the learner to recover a rule that achieves perfect classification from history alone at prediction time. 

\section{ForeClassNet Architecture}

In this section, we present ForeClassNet and its main architectural components. ForeClassNet comprises a forecasting module and a classification module. For each stochastic forward pass, the forecasting module generates a future trajectory, which is concatenated with the observed time series and passed to the classification module. The forecasting module directly predicts all $k$ future time steps in a single forward pass, rather than recursively generating forecasts one step at a time. We use Monte Carlo dropout \cite{gal2016dropout}, with dropout active during both training and inference, to generate multiple forecast trajectories and their associated class-probability vectors. Averaging these vectors provides a Monte Carlo approximation to the posterior predictive class probabilities under the dropout approximation. The averaged class probabilities are then used to make the final classification decision. ForeClassNet also incorporates Boltzmann convolutional layers, which learn a softmax-weighted combination of convolutional outputs associated with different filter lengths.

\begin{figure}[h!] 
    \centering
    \includegraphics[width=\textwidth]{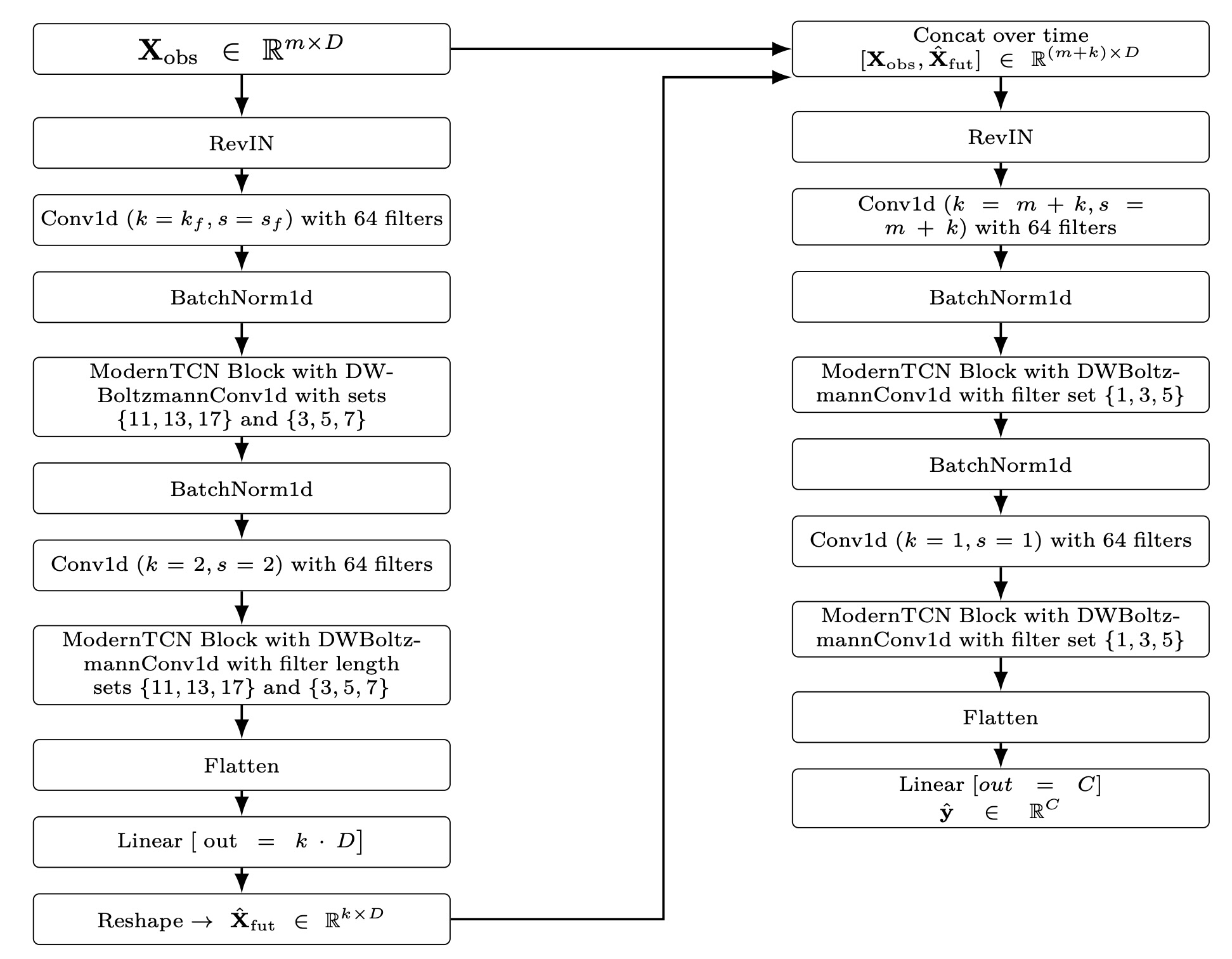} 
    \caption{Diagram of the proposed ForeClassNet architecture, where DW denotes a depthwise convolutional layer. Further details on ModernTCN blocks are provided in \cite{luo2024moderntcn}. Monte Carlo dropout is applied throughout the network with a dropout probability of 0.1. Here, $L_{\text{obs}} = m$ denotes the observed time series length, $k_{f} = \min\{16, L_{\text{obs}}\}$, and $s_{f} = \min\{8, k_{f}\}$.  }
    \label{fig:final_model} 
\end{figure}

We first define Reversible Instance Normalization (RevIN) \cite{kim2021reversible}. For a multivariate time series $x_{i}$ with $d_{\text{feat}}$ variables, the normalization and denormalization steps of RevIN are defined as follows: 
\begin{align*}
    \mu_{i,d} = \frac{1}{m} \sum_{t=1}^{m} x_{i,t,d}, \qquad
    \sigma_{i,d} = \sqrt{\frac{1}{m} \sum_{t=1}^{m} (x_{i,t,d} - \mu_{i,d})^{2} + \epsilon}, \qquad 
    \tilde{x}_{i,t,d}= 
       \frac{x_{i,t,d} - \mu_{i,d}}{\sigma_{i,d}}, 
\end{align*}
where $\epsilon$ is a small numerical offset. The forecasting module uses the normalized observations to produce forecasts $\hat{\tilde{x}}_{i,m+r,d}$ on the normalized scale. These forecasts are transformed back to the original scale using 
\begin{equation*}
    \hat{x}_{i,m+r,d} = \sigma_{i,d}\hat{\tilde{x}}_{i,m+r,d} + \mu_{i,d}, \qquad r=1, \dots, k.
\end{equation*}
Figure \ref{fig:final_model} provides an overview of the proposed ForeClassNet architecture. 

The forecasting and classification modules reflect the distinction in Definition 1 between the process generating future observations and the rule determining the class label. Result 1 identifies the excess classification error relative to the history-only Bayes risk as the component that improved learning can reduce. Accordingly, ForeClassNet uses realized future observations as additional supervision during training, with the aim of reducing this excess error. Proposition 1 provides a finite-sample example illustrating how future observations used during training can improve predictions made from observed histories alone, and Result 2 characterizes how discrepancies in the generated forecast distribution contribute to an upper bound on classification error. 

In the original ModernTCN architecture, the model comprises several parallel blocks, each containing two branches: one branch captures short-range receptive fields, whereas the other captures long-range receptive fields. Although Monte Carlo dropout provides an approximate Bayesian treatment of network parameters, the convolutional filter lengths within these branches remain fixed in advance. This may be restrictive because the temporal scales most relevant for forecasting and classification can vary across data sets. To reduce reliance on a single predetermined filter length within each branch, we introduce Boltzmann convolutions (BCs), which learn a softmax-weighted combination of convolutional outputs associated with different candidate filter lengths. Let $\mathcal{K}=\{\kappa_{1},\ldots,\kappa_{J}\}$ denote a set of candidate filter lengths, and let $p_{j}$ be the learned logit associated with filter length $\kappa_{j}$. The combination weights are defined using a softmax with temperature $T>0$:
\begin{equation*}
    \pi_{j} = \frac{\exp\{ \frac{p_{j}}{T} \}}{\sum\limits_{r=1}^{J} \exp\{\frac{p_{r}}{T} \} }. 
\end{equation*}

Let $C_{j}$ denote the convolutional layer associated with filter length $\kappa_{j}$. The Boltzmann convolution combines the outputs of these layers as 
\begin{equation*}
    \operatorname{BC}(x) = \sum\limits_{j=1}^{J} \pi_{j}(C_{j}*x),
\end{equation*}
where $*$ denotes convolution. In our models, we set $T=1$ and use candidate filter length sets $\{ 3,5,7 \}$ and $\{ 11,13,17 \}$. 

We also consider ForeClassNet Dynamic, a variant in which Boltzmann convolutions are replaced by input-dependent dynamic convolutions, following the attention-based approach of \cite{chen2020dynamic}. The dynamic variant computes convolutional combination weights separately for each input, rather than using the learned weights shared across inputs in BCs. The remaining forecasting and classification architecture is unchanged. 

The standard ForeClassNet model and the dynamic ForeClassNet model are trained by minimizing the following joint loss function: 
\begin{align*}
    \text{Loss}(\boldsymbol{x}_{*}^{(i)}, \hat{\boldsymbol{x}}_{*}^{(i)}, \boldsymbol{y}^{(i)}, \hat{\boldsymbol{y}}^{(i)})  &= \text{MSE}(\boldsymbol{x}_{*}^{(i)},\hat{\boldsymbol{x}}_{*}^{(i)} ) + \text{CCE}(\boldsymbol{y}^{(i)},\hat{\boldsymbol{y}}^{(i)} ) \\ 
    & =  \frac{1}{k d_{\text{feat}}} \sum\limits_{t=1}^{k} \sum\limits_{d=1}^{d_{\text{feat}}} \left(x_{*,t,d}^{(i)} - \hat{x}_{*,t,d}^{(i)}\right)^{2} - \sum\limits_{j=1}^{L} y^{(i)}_{j} \log(\hat{y}^{(i)}_{j}),
\end{align*}
where $\boldsymbol{y}^{(i)}$ is the one-hot-encoded vector of the true class label for example $i$. The vector $\hat{\boldsymbol{y}}^{(i)}$ contains the predicted probabilities for the $L$ classes for example $i$. 

The choice of a joint forecasting and classification objective is further motivated by Corollary 1 and Result 3. Corollary 1 relates excess classification error to classification cross-entropy, while Result 3 extends this analysis to a joint forecasting and classification objective under its stated assumptions. We adopt the unweighted sum of forecasting MSE and classification cross-entropy as a simple baseline objective, while investigating the effect of alternative approaches to balancing the two tasks through the constrained training objectives introduced below that minimize the classification cross-entropy subject to a bound on forecasting loss. 

The first retains mean squared error as the forecasting criterion, whereas the second uses the energy score to assess the predictive distribution of future trajectories. For these variants, classification cross-entropy is calculated from the class probabilities averaged across stochastic forward passes. Writing $C(\boldsymbol{\theta})$ for the empirical classification cross-entropy and $F_{r}(\boldsymbol{\theta})$ for the empirical forecasting criterion, these formulations are 
\begin{equation*}
    \min\limits_{\boldsymbol{\theta}} C(\boldsymbol{\theta}) \text{ subject to } F_{r}(\boldsymbol{\theta}) \leq b_{r}, r\in \{ MSE, ES \},
\end{equation*}
where $\boldsymbol{\theta}$ denotes the model parameters. Their corresponding Lagrangian objectives are 
\begin{align*}
    \mathcal{L}_{MSE}(\boldsymbol{\theta}, \lambda_{MSE}) &= C(\boldsymbol{\theta}) + \lambda_{MSE}(\frac{F_{MSE}(\boldsymbol{\theta})}{b_{MSE}}-1)\\
    \mathcal{L}_{ES}(\boldsymbol{\theta} , \lambda_{ES}) &= C(\boldsymbol{\theta}) + \lambda_{ES}(\frac{F_{ES}(\boldsymbol{\theta})}{b_{ES}}-1),
\end{align*}
with nonnegative multipliers updated by dual ascent during training and $b_{MSE},b_{ES}>0$. Each bound is set using a bootstrap upper confidence bound on the corresponding validation loss following forecasting-only pretraining. For a predictive distribution $P$ and observed future trajectory $\boldsymbol{x}_{*}$, the energy score is 
\begin{equation*}
    ES(P,\boldsymbol{x}_{*}) = \mathbb{E}[||Z-x_{*}||_{2}] - \frac{1}{2} \mathbb{E}[||Z-Z'||_{2}], 
\end{equation*}
where $Z$ and $Z'$ are independent draws from $P$, and the expectations are approximated using Monte Carlo samples. The energy score thus provides a measure of the quality of predictive distributions \cite{gneiting2007strictly}. Together, the two alternatives examine adaptive weighting while retaining squared-error forecasting and the additional effect of using a distributional forecasting criterion. 

\section{Experiments}

We implement ForeClassNet and the competing models in Python \cite{10.5555/1593511}, using PyTorch \cite{paszke2019pytorch} for the neural network models. The experiments are conducted on a computer cluster where computational hardware is allocated to individual computing jobs according to availability. Consequently, different experiments and training runs are executed on different hardware configurations. To ensure that computational costs are reported appropriately, the corresponding tables identify the hardware used and report training and inference times separately for each hardware configuration. The code to reproduce our experiments is available at \url{https://github.com/DanielCoulson/Foreclassing-experiments}. In Sections 4.1 and 4.2, we report the following classification performance and uncertainty metrics, while finance-specific metrics are defined and reported in Section 4.3. In each case, TP, TN, FP, and FN denote the numbers of true positives, true negatives, false positives, and false negatives, respectively.  
Accuracy is the proportion of predictions that are correct:
\begin{equation*}
    \text{Accuracy} = \frac{TP + TN}{TP + TN + FP + FN}.
\end{equation*}
The area under the receiver operating characteristic curve (AUROC) is the probability that a classifier assigns a higher score to a randomly chosen positive example than to a randomly chosen negative example. Let $s^{+}$ denote the score assigned to a positive example and $s^{-}$ denote the score assigned to a negative example. Then,
\begin{equation*}
    \text{AUROC} = \mathbb{P}(s^{+}>s^{-}) + \frac{1}{2}\mathbb{P}(s^{+} = s^{-}).
\end{equation*} 
We also report the average precision score (AP), which summarizes the precision--recall trade-off by averaging precision over different recall levels. Intuitively, AP measures how well the model maintains correct positive predictions as we consider increasingly many positive classifications. Let $K$ denote the number of operating points on the precision--recall curve with the initial recall set to zero. Then,
\begin{equation*}
    \text{AP} = \sum\limits_{k=1}^{K}(\frac{TP_{k}}{TP_{k} + FN_{k}} - \frac{TP_{k-1}}{TP_{k-1}+FN_{k-1}})(\frac{TP_{k}}{TP_{k} + FP_{k}}). 
\end{equation*}
We also report the F1 score, which is the harmonic mean of precision and recall. For the positive class it is given by 
\begin{equation*}
    \text{F1} = \frac{2TP}{2TP+FP+FN}. 
\end{equation*}
To evaluate probabilistic predictions, we report the Brier score, negative log-likelihood (NLL), and expected calibration error (ECE). Let $N$ denote the number of test observations, $y_{i} \in \{0,1\}$ the true class label, and $p_{i}$ the predicted probability of class 1 for observation $i$. 
The Brier score measures the mean squared error of the predicted probabilities: 
\begin{equation*}
    \text{Brier} = \frac{1}{N}\sum\limits_{i=1}^{N} (p_{i} - y_{i})^{2}, 
\end{equation*}
with lower values indicating more accurate probabilistic predictions. 
NLL evaluates the probability assigned to the observed class labels: 
\begin{equation*}
    \text{NLL} = - \frac{1}{N}\sum\limits_{i=1}^{N}(y_{i} \log(p_{i}) + (1-y_{i})\log(1-p_{i})), 
\end{equation*}
where lower NLL indicates better probabilistic predictions, with larger penalties assigned to confident incorrect predictions. 
ECE measures the discrepancy between predicted probabilities and observed event frequencies. We divide the predicted class-1 probabilities into $B=10$ equally spaced bins. Let $\mathcal{B}_{b}$ denote the set of observations in bin $b$, with $n_{b} = |\mathcal{B}_{b}|$. We calculate 
\begin{equation*}
    \text{ECE} = \sum\limits_{b=1}^{B}\frac{n_{b}}{N}|\frac{1}{n_{b}} \sum\limits_{i \in \mathcal{B}_{b}}p_{i} - \frac{1}{n_{b}}\sum\limits_{i \in \mathcal{B}_{b}} y_{i}|
\end{equation*}
where empty bins contribute zero and lower ECE indicates closer agreement between predicted probabilities and observed event frequencies. 

We compare the standard ForeClassNet model and ForeClassNet Dynamic with ForeClassNet MSE and ForeClassNet Energy, which use the alternative training objectives described in Section 3. We also consider ModernTCN \cite{luo2024moderntcn} and a Forecast-then-Classify version of ModernTCN. The ModernTCN networks use two stages, each containing a ModernTCN block with 64 feature channels. For the Forecast-then-Classify baseline, separate ModernTCN networks are used for forecasting and classification. As a further benchmark, we include MultiRocketHydra, which is identified as a leading time series classifier in the large-scale review of \cite{middlehurst2024bake}. However, in Section 4.1 we instead use a multivariate ROCKET classifier \cite{dempster2020rocket}, with 10,000 random convolutional kernels of lengths 2, 3, and 4, followed by ridge classification, due to the short observation windows. We additionally consider two Transformer models: a standalone Transformer classifier and a Transformer Forecast-then-Classify model. 

The standalone Transformer classifier consists of two Transformer encoder layers, each with four attention heads, a model dimension of 64, and a feedforward dimension of 128. The architecture uses sinusoidal positional encodings, GELU activations, dropout with probability 0.1, and Reversible Instance Normalization (RevIN). The encoded representations are averaged across the temporal dimension and passed through a linear classification head. The Transformer Forecast-then-Classify model uses a forecasting network with two encoder layers and two decoder layers, followed by a separate Transformer classifier with two encoder layers. The forecasting and classification components use the same model dimension, number of attention heads, feedforward dimension, dropout probability, and normalization scheme as the standalone Transformer classifier. The forecasting decoder uses learned query embeddings for future time steps, allowing it to predict the entire forecasting horizon directly from observed history. The resulting forecasts are concatenated with the observed history and provided as input to the classification network. 

All neural network models are trained using the Adam optimizer of \cite{kingma2014adam} with a learning rate of $10^{-3}$, weight decay of $10^{-4}$ and a batch size of 512. The standard ForeClassNet and ForeClassNet Dynamic models are trained jointly for 50 epochs by minimizing the sum of forecasting mean squared error and classification cross-entropy. The standalone ModernTCN and Transformer classifiers are also trained for 50 epochs using classification cross-entropy. For the ModernTCN and Transformer Forecast-then-Classify models, the forecasting networks are first trained for 20 epochs using mean squared error, followed by 50 epochs of training for their respective classifiers. ForeClassNet MSE and ForeClassNet Energy undergo 20 epochs of forecasting-only pretraining, followed by 50 epochs of constrained joint optimization. For these constrained variants, 20\% of the training data is reserved for validation, and the forecasting-loss bounds are determined from 95\% bootstrap upper confidence bounds using 2,000 bootstrap replicates. Constrained training uses eight Monte Carlo samples per observation, with 32 samples used to estimate the forecasting criterion for the constraint bound. The Lagrange multipliers are initialized at zero and updated by dual ascent with a learning rate of $0.01$. 

For the ForeClassNet models, the default Monte Carlo dropout probability is 0.1, and the Boltzmann convolution temperature is set to $T=1$. ForeClassNet Dynamic uses input-dependent convolutional weights obtained from an attention mechanism with a channel-reduced factor of four and a minimum hidden dimension of eight. Each neural network model is trained five times using the fixed random seeds 0,1,2,3, and 4, with model parameters reinitialized for each run. For each ForeClassNet variant, class probabilities at inference are obtained by averaging the predictions from 100 Monte Carlo dropout forward passes. Performance metrics are calculated separately for each seed and then averaged across the five runs. In Sections 4.4 and 4.5, we consider sensitivity and ablation studies of various components of the ForeClassNet model. 

To assess statistical significance, we use a moving-block bootstrap with 5,000 replicates and a block length equal to the forecasting horizon. For each comparison, the same temporal blocks are resampled for the competing models, and the resulting performance differences are used to estimate their sampling distribution. For the weather experiments in Section 4.2, resampling is performed separately within each test year, and the resulting differences are averaged across years. We construct an estimate of the null distribution by subtracting the observed performance metric difference from the bootstrap differences. We then calculate the $p$-value as the proportion of null-centred bootstrap differences at least as extreme as the observed difference specified by the alternative hypothesis. To account for multiple testing, we use a fixed-sequence testing procedure for AP followed by AUROC and Brier score followed by NLL. 

\subsection{Power transformer monitoring}

Power transformers are critical components of national electrical grids. Therefore, monitoring transformers is essential for ensuring reliable operation. A commonly used indicator of transformer health is oil temperature. In this experiment, we use the electricity transformer dataset from \cite{zhou2021informer}. We consider a single transformer and determine whether the oil temperature will exceed a threshold of $13^\circ$C. The data consist of multivariate time series observations. We observe the first hour of a transformer's measurements and aim to predict whether the oil temperature will exceed the specified threshold during the subsequent 12 hours. The dataset contains seven variables: high useful fuel load, high useless fuel load, middle useful fuel load, middle useless fuel load, low useful fuel load, low useless fuel load, and oil temperature. The data are constructed into rolling windows. At each 15-minute step, we use the previous 60 minutes, corresponding to four time steps, as input and predict whether the oil temperature will exceed the threshold within the next 12 hours. This procedure yields 55,693 training windows, which are downsampled to 42,134 to create a balanced training set, and a temporally out-of-distribution test set of 13,885 windows, where all timestamps in the test set occur after those in the training set. This is an imbalanced classification problem, with only $13.5\%$ of the time series in the test set exceeding the temperature threshold. Therefore, we are primarily interested in the classifier's performance on the positive minority class, corresponding to cases in which the temperature exceeds the threshold. The performance results are provided in Tables 1 and 2 with computing time reported in Table 3. 

\begin{table*}[t]
\centering
\caption{Classification performance on the Oil dataset. Results are averaged across five training seeds.}
\label{tab:oil_performance}
\begin{tabular}{lcccc}
\toprule
Model & Accuracy  & F1 & AUROC & AP \\
\midrule
ModernTCN
& 0.847  & 0.590 & 0.928 & 0.742 \\

ModernTCN Forecast-then-Classify
& 0.875  & 0.617 & 0.922 & 0.726 \\

ForeClassNet
& 0.897  & 0.621 & 0.929 & 0.754 \\

Transformer Classifier
& 0.557  & 0.235 & 0.554 & 0.157 \\

Transformer Forecast-then-Classify
& 0.525  & 0.241 & 0.564 & 0.159 \\

ForeClassNet Dynamic
& 0.877  & 0.618 & 0.932 & \textbf{0.762} \\

ForeClassNet MSE
& \textbf{0.899}  & \textbf{0.642} & 0.931 & 0.749 \\

ForeClassNet Energy
& 0.893  & \textbf{0.642} & \textbf{0.933} & 0.751 \\

ROCKET
& 0.882  & 0.548 & 0.883 & 0.598 \\
\bottomrule
\end{tabular}
\end{table*}

\begin{table*}[t]
\centering
\caption{Uncertainty performance on the Oil dataset. Results are averaged across five training seeds.}
\label{tab:oil_uncertainty}
\begin{tabular}{lccc}
\toprule
Model & Brier & NLL  & ECE   \\
\midrule
ModernTCN
& 0.1086 & 0.3429 & 0.1215 \\

ModernTCN Forecast-then-Classify
& 0.0887 & 0.2803 & 0.0840 \\

ForeClassNet
& 0.0738 & 0.2388 & 0.0685  \\

Transformer Classifier
& 0.2472 & 0.6884 & 0.3529 \\

Transformer Forecast-then-Classify
& 0.2530 & 0.7004 & 0.3610  \\

ForeClassNet Dynamic
& 0.0838 & 0.2629 & 0.0900  \\

ForeClassNet MSE
& \textbf{0.0709} & \textbf{0.2263} & \textbf{0.0538}  \\

ForeClassNet Energy
& 0.0743 & 0.2352 & 0.0674 \\
\bottomrule
\end{tabular}
\end{table*}

\begin{table*}[t]
\centering
\caption{Computational cost on the Oil dataset. Times are averaged only across
runs executed on the same recorded hardware.}
\label{tab:oil_computational_cost}
\resizebox{\linewidth}{!}{%
\begin{tabular}{llrrr}
\toprule
Model & Hardware & Runs & Training (min) & Inference (ms/example) \\
\midrule
ModernTCN
& GTX TITAN X & 5 & 2.52 & 0.022 \\

ModernTCN Forecast-then-Classify
& GTX TITAN X & 5 & 3.13 & 0.033 \\

ForeClassNet
& GTX TITAN X & 2 & 7.42 & 4.817 \\

ForeClassNet
& TITAN X (Pascal) & 3 & 5.21 & 4.188 \\

Transformer Classifier
& TITAN Xp & 5 & 1.65 & 0.013 \\

Transformer Forecast-then-Classify
& TITAN Xp & 5 & 3.17 & 0.049 \\

ForeClassNet Dynamic
& TITAN Xp & 5 & 4.97 & 3.689 \\

ForeClassNet MSE
& GTX 1080 Ti & 4 & 34.61 & 4.309 \\

ForeClassNet MSE
& Tesla T4 & 1 & 23.79 & 3.421 \\

ForeClassNet Energy
& GTX 1080 Ti & 5 & 36.30 & 4.263 \\

ROCKET
& CPU & 5 & 39.60 & 1.108 \\
\bottomrule
\end{tabular}%
}
\end{table*}

Table~\ref{tab:oil_performance} shows that the two primary ForeClassNet architectures perform well relative to competing methods. ForeClassNet achieves an AP of 0.754 and AUROC of 0.929, while ForeClassNet Dynamic achieves an AP of 0.762 and AUROC of 0.932. Both architectures exceed the corresponding results of the two ModernTCN models, with ForeClassNet Dynamic achieving the highest AP among all evaluated models. The AP improvement of the primary ForeClassNet models compared to the ModernTCN models is statistically significant with a p-value of $0.037$, while the improvement in AUROC is marginally insignificant with a p-value of $0.067$. The alternative ForeClassNet variants also perform competitively: ForeClassNet MSE achieves the highest accuracy of 0.899 and is tied for the highest F1 score of 0.642, while ForeClassNet Energy achieves the highest AUROC of 0.933. Both primary ForeClassNet architectures also achieve higher AUROC and AP than ROCKET and the two transformer models. 

Table~\ref{tab:oil_uncertainty} indicates the primary ForeClassNet architectures also provide improvements in probabilistic prediction. ForeClassNet achieves a Brier score of 0.0738 and NLL of 0.2388, while ForeClassNet Dynamic achieves corresponding values of 0.0838 and 0.2629, compared with the lowest Brier score of 0.0887 and NLL of 0.2803 among the ModernTCN models. The combined comparison of the primary architectures with the ModernTCN models shows statistically significant reductions in both Brier score with a p-value of 0.030 and NLL with a p-value of 0.027. Among the alternative variants, ForeClassNet MSE achieves the lowest Brier score of 0.0709, NLL of 0.2263, and ECE of 0.0538 of all evaluated models. Both primary ForeClassNet architectures achieve lower Brier scores, NLL, and ECE than the two transformer models. 

We repeat a similar experiment for a separate power transformer with a temperature threshold of $30^\circ\text{C}$. This results in a test set with 43.5\% of examples in class 0 and 56.5\% in class 1. Accordingly, we report class 0 AP and class 1 AP, since our objective is to predict when transformers will exceed the temperature threshold (class 1) while also avoiding overfitting to the majority class.

\begin{table*}[t]
\centering
\caption{Classification performance on Oil Example 2. Results are averaged across five training seeds. }
\label{tab:oil2_performance}
\begin{tabular}{lccccc}
\toprule
Model & Accuracy  & F1 & AUROC & AP Class 0 & AP Class 1 \\
\midrule
ModernTCN
& 0.8636  & 0.8810 & 0.9483 & 0.9276 & 0.9650 \\

ModernTCN Forecast-then-Classify
& 0.8670  & 0.8818 & 0.9505 & 0.9327 & 0.9658 \\

ForeClassNet
& 0.8665  & 0.8842 & 0.9562 & 0.9426 & 0.9692 \\

Transformer Classifier
& 0.6058 & 0.6373 & 0.6460 & 0.6173 & 0.6628 \\

Transformer Forecast-then-Classify
& 0.6164  & 0.6593 & 0.6453 & 0.6108 & 0.6594 \\

ForeClassNet Dynamic
& 0.8745  & 0.8869 & 0.9584 & 0.9456 & 0.9707 \\

ForeClassNet MSE
& 0.8794  & 0.8965 & \textbf{0.9604} & \textbf{0.9481} & 0.9718 \\

ForeClassNet Energy
& \textbf{0.8805}  & \textbf{0.8972} & 0.9602 & 0.9476 & \textbf{0.9719} \\

ROCKET
& 0.8636  & 0.8798 & 0.9304 & 0.9146 & 0.9337 \\
\bottomrule
\end{tabular}
\end{table*}

\begin{table*}[t]
\centering
\caption{Uncertainty and calibration performance on Oil Example 2. Results are averaged across five training seeds.}
\label{tab:oil2_uncertainty}
\begin{tabular}{lccc}
\toprule
Model & Brier  & NLL & ECE \\
\midrule
ModernTCN
& 0.1014 & 0.3567 & 0.0698  \\

ModernTCN Forecast-then-Classify
& 0.0942 & 0.3001 & 0.0523  \\

ForeClassNet
& 0.0931 & 0.2944 & 0.0714  \\

Transformer Classifier
& 0.2364 & 0.6684 & 0.0937  \\

Transformer Forecast-then-Classify
& 0.2351 & 0.6673 & 0.0815 \\

ForeClassNet Dynamic
& \textbf{0.0851} & \textbf{0.2674} & \textbf{0.0458}\\

ForeClassNet MSE
& 0.0889 & 0.2835 & 0.0672  \\

ForeClassNet Energy
& 0.0885 & 0.2808 & 0.0652  \\
\bottomrule
\end{tabular}
\end{table*}

\begin{table*}[t]
\centering
\caption{Computational cost on Oil Example 2. Times are averaged only across
runs executed on the same recorded hardware. }
\label{tab:oil2_computational_cost}
\resizebox{\linewidth}{!}{%
\begin{tabular}{llrrr}
\toprule
Model & Hardware & Runs & Training (min) & Inference (ms/example) \\
\midrule
ModernTCN
& NVIDIA L4 & 2 & 1.05 & 0.008 \\

ModernTCN
& NVIDIA RTX A5000 & 2 & 0.99 & 0.008 \\

ModernTCN
& Tesla V100-SXM3-32GB-H & 1 & 2.64 & 0.010 \\

ModernTCN Forecast-then-Classify
& NVIDIA RTX A5000 & 3 & 2.25 & 0.025 \\

ModernTCN Forecast-then-Classify
& Tesla V100-SXM3-32GB-H & 2 & 3.42 & 0.023 \\

ForeClassNet
& NVIDIA RTX A5000 & 5 & 4.11 & 2.581 \\

Transformer Classifier
& NVIDIA RTX A5000 & 5 & 3.05 & 0.008 \\

Transformer Forecast-then-Classify
& NVIDIA RTX A5500 & 5 & 3.58 & 0.027 \\

ForeClassNet Dynamic
& NVIDIA GeForce RTX 2080 Ti & 5 & 5.68 & 3.683 \\

ForeClassNet MSE
& NVIDIA GeForce RTX 2080 Ti & 5 & 21.63 & 2.629 \\

ForeClassNet Energy
& NVIDIA GeForce RTX 2080 Ti & 5 & 25.26 & 3.017 \\

ROCKET
& CPU & 5 & 46.10 & 1.160 \\
\bottomrule
\end{tabular}%
}
\end{table*}

Table~\ref{tab:oil2_performance} shows that the two primary ForeClassNet architectures achieve higher AUROC and average precision for both classes than the two ModernTCN models. ForeClassNet achieves an AUROC of 0.9562, class-0 AP of 0.9426, and class-1 AP of 0.9692, while ForeClassNet Dynamic achieves corresponding values of 0.9584, 0.9456, and 0.9707. For comparison, the ModernTCN Forecast-then-Classify baseline achieves an AUROC of 0.9505, class-0 AP of 0.9327, and class-1 AP of 0.9658. The improvements in class-1 AP and AUROC of the primary ForeClassNet models compared to the ModernTCN models are statistically significant with p-values of 0.020 and 0.018 respectively. The alternative training objectives give further improvements in some of the metrics, with ForeClassNet MSE achieving the highest AUROC of 0.9604 and class-0 AP of 0.9481, and ForeClassNet Energy achieving the highest class-1 AP of 0.9719, accuracy of 0.8805, and an F1 score of 0.8972. Both primary ForeClassNet architectures achieve higher AUROC and average precision for both classes than ROCKET and the two transformer models.  

Table~\ref{tab:oil2_uncertainty} shows that ForeClassNet Dynamic achieves the lowest Brier score of 0.0851, NLL of 0.2674, and ECE of 0.0458. The standard ForeClassNet model also achieves lower Brier and NLL scores than either ModernTCN baseline, but with slightly higher ECE on average. The reduction in Brier score and NLL of the primary ForeClassNet models compared to the ModernTCN models is statistically significant with p-values of 0.031 and 0.023 respectively. The primary ForeClassNet architectures outperform both Transformer models across all three metrics. 

Tables~\ref{tab:oil_computational_cost} and \ref{tab:oil2_computational_cost} show that the two primary ForeClassNet architectures generally require more computation than the ModernTCN models. Nevertheless, their absolute computational costs remain relatively modest, with training times of approximately 4--8 minutes and inference times of approximately 2.6--4.8 milliseconds per example, despite using 100 Monte Carlo dropout forward passes. 

Taken together, the two oil-temperature experiments provide evidence that the primary ForeClassNet architectures can improve classification discrimination and probabilistic prediction relative to competing methods. These improvements are observed across two different temperature thresholds and test-set class distributions. 

\subsection{Extreme Weather}

In this experiment, we examine extreme weather conditions in Shanghai, China. The weather data comprise five features: wind speed at 10 m above ground level (m/s), relative humidity at 2 m ($\%$), daily maximum air temperature at 2 m ($^\circ\mathrm{C}$), daily minimum air temperature at 2 m ($^\circ\mathrm{C}$), and total bias-corrected precipitation (mm/day). The data were obtained from the National Aeronautics and Space Administration (NASA) Langley Research Center's Prediction of Worldwide Energy Resources (POWER) project, funded through the NASA Earth Science Division. Given the previous 30 daily observations of these variables, the task is to predict whether an extreme temperature warning should be issued over the subsequent seven-day period. We define an extreme temperature event as a day on which the minimum temperature falls below the 5th percentile of all minimum temperatures, or the maximum temperature exceeds the 95th percentile of all maximum temperatures. As extreme weather events constitute the minority class, we are particularly interested in classification performance for class 1, while also reporting average precision for class 0. For each test year, each model is trained using all observations prior to that year and evaluated on the held-out year. The models are reinitialized for each test year, and the procedure is repeated using five training seeds. Results are averaged across seeds within each year and then across test years. Classification and probabilistic results are provided in Tables~\ref{tab:shanghai_performance} and \ref{tab:shanghai_uncertainty}, respectively. 

\begin{table*}[t]
\centering
\caption{Classification performance on the Shanghai extreme-weather dataset
over the 2014--2021 test years. Results are averaged across five training seeds
for each year and then across years.}
\label{tab:shanghai_performance}
\begin{tabular}{lccccc}
\toprule
Model & Accuracy & F1 & AUROC & AP Class 0 & AP Class 1 \\
\midrule

ModernTCN
& 0.8335 & 0.5639 & 0.8659 & 0.9606 & 0.6131 \\

ModernTCN Forecast-then-Classify
& 0.8365 & 0.5721 & 0.8728 & 0.9632 & 0.6379 \\

ForeClassNet
& \textbf{0.8512} & 0.5807 & \textbf{0.8865}
& \textbf{0.9686} & \textbf{0.6793} \\

Transformer Classifier
& 0.7804 & 0.4226 & 0.7490 & 0.9177 & 0.4298 \\

Transformer Forecast-then-Classify
& 0.7786 & 0.4368 & 0.7670 & 0.9268 & 0.4518 \\

ForeClassNet Dynamic
& 0.8496 & \textbf{0.5938} & 0.8843 & 0.9667 & 0.6670 \\

ForeClassNet MSE
& 0.8334 & 0.5549 & 0.8556 & 0.9575 & 0.5984 \\

ForeClassNet Energy
& 0.8350 & 0.5559 & 0.8558 & 0.9584 & 0.5904 \\

MRHydra
& 0.7717 & 0.4181 & 0.6438 & 0.8504 & 0.3039 \\

\bottomrule
\end{tabular}
\end{table*}

\begin{table*}[t]
\centering
\caption{Uncertainty and calibration performance on the Shanghai
extreme-weather dataset over the 2014--2021 test years. Results are averaged
across five training seeds for each year and then across years.}
\label{tab:shanghai_uncertainty}
\begin{tabular}{lccc}
\toprule
Model & Brier & NLL & ECE \\
\midrule

ModernTCN
& 0.1176 & 0.3779 & 0.0781 \\

ModernTCN Forecast-then-Classify
& 0.1136 & 0.3614 & 0.0755 \\

ForeClassNet
& \textbf{0.1030} & \textbf{0.3283} & 0.0735 \\

Transformer Classifier
& 0.1837 & 0.9075 & 0.1733 \\

Transformer Forecast-then-Classify
& 0.1878 & 0.9250 & 0.1797 \\

ForeClassNet Dynamic
& 0.1043 & 0.3316 & \textbf{0.0713} \\

ForeClassNet MSE
& 0.1183 & 0.3769 & 0.0743 \\

ForeClassNet Energy
& 0.1180 & 0.3756 & 0.0714 \\

\bottomrule
\end{tabular}
\end{table*}

\begin{table*}[t]
\centering
\caption{Computational cost on the Shanghai extreme-weather dataset
over the 2014--2021 test years. Neural-network timings are averaged
over runs executed on the NVIDIA GeForce RTX 2080 Ti.
MRHydra timings are averaged over 40 CPU runs.}
\label{tab:shanghai_computational_cost}
\resizebox{\linewidth}{!}{%
\begin{tabular}{llrrr}
\toprule
Model & Hardware & Runs & Training (min) & Inference (ms/example) \\
\midrule

ModernTCN
& NVIDIA GeForce RTX 2080 Ti & 10 & 0.39 & 0.038 \\

ModernTCN Forecast-then-Classify
& NVIDIA GeForce RTX 2080 Ti & 14 & 0.50 & 0.053 \\

ForeClassNet
& NVIDIA GeForce RTX 2080 Ti & 11 & 1.11 & 3.752 \\

Transformer Classifier
& NVIDIA GeForce RTX 2080 Ti & 10 & 0.71 & 0.053 \\

Transformer Forecast-then-Classify
& NVIDIA GeForce RTX 2080 Ti & 5 & 0.90 & 0.046 \\

ForeClassNet Dynamic
& NVIDIA GeForce RTX 2080 Ti & 7 & 1.26 & 4.649 \\

ForeClassNet MSE
& NVIDIA GeForce RTX 2080 Ti & 9 & 5.55 & 3.970 \\

ForeClassNet Energy
& NVIDIA GeForce RTX 2080 Ti & 7 & 5.84 & 4.386 \\

MRHydra
& CPU & 40 & 5.17 & 5.985 \\

\bottomrule
\end{tabular}%
}
\end{table*}

Table~\ref{tab:shanghai_performance} shows that the two primary ForeClassNet architectures achieve higher accuracy, F1 score, AUROC and average precision for both classes than the two ModernTCN models. ForeClassNet achieves an AUROC of 0.8865, class-0 AP of 0.9686, and class-1 AP of 0.6793, while ForeClassNet Dynamic achieves corresponding values of 0.8843, 0.9667, and 0.6670. ForeClassNet also achieves the highest accuracy of 0.8512, while ForeClassNet Dynamic achieves the highest F1 score of 0.5938. The standard ForeClassNet achieves the highest class-1 AP among all evaluated models. The improvements in class-1 AP and AUROC of the primary ForeClassNet models compared to the ModernTCN models are statistically significant, with p-values of 0.0020 and 0.011, respectively. Both primary architectures also outperform the Transformer and MRHydra models on these measures. In contrast, the ForeClassNet MSE and Energy variants achieve lower discrimination performance in this experiment.

Table~\ref{tab:shanghai_uncertainty} shows that both primary ForeClassNet architectures achieve lower Brier scores and NLL than the two ModernTCN models. ForeClassNet achieves a Brier score of 0.1030, NLL of 0.3283, and ECE of 0.0735, while ForeClassNet Dynamic achieves corresponding values of 0.1043, 0.3316, and 0.0713. The standard ForeClassNet achieves the lowest Brier score and NLL, while ForeClassNet Dynamic achieves the lowest ECE. The improvements of the primary ForeClassNet architectures against the ModernTCN models for both Brier score and NLL are statistically significant, with p-values of 0.0004 and 0.0006, respectively. The primary ForeClassNet architectures also achieve lower Brier scores, NLL, and ECE than the Transformer models. We repeat the extreme-weather experiment for New York City and Tokyo. 

\begin{table*}[t]
\centering
\caption{Classification performance on the New York City extreme-weather
dataset over the 2014--2021 test years. Results are averaged across five
training seeds for each year and then across years.}
\label{tab:nyc_performance}
\begin{tabular}{lccccc}
\toprule
Model & Accuracy & F1 & AUROC & AP Class 0 & AP Class 1 \\
\midrule

ModernTCN
& 0.8241 & 0.6192 & 0.8742 & 0.9521 & 0.6985 \\

ModernTCN Forecast-then-Classify
& 0.8242 & 0.6227 & 0.8700 & 0.9507 & 0.6925 \\

ForeClassNet
& \textbf{0.8500} & \textbf{0.6665}
& 0.8921 & 0.9597 & 0.7441 \\

Transformer Classifier
& 0.7254 & 0.4227 & 0.6982 & 0.8626 & 0.4592 \\

Transformer Forecast-then-Classify
& 0.7238 & 0.4021 & 0.6959 & 0.8609 & 0.4521 \\

ForeClassNet Dynamic
& 0.8458 & 0.6603
& \textbf{0.8940} & \textbf{0.9610} & \textbf{0.7499} \\

ForeClassNet MSE
& 0.8235 & 0.6329 & 0.8629 & 0.9433 & 0.6934 \\

ForeClassNet Energy
& 0.8157 & 0.6192 & 0.8568 & 0.9420 & 0.6729 \\

MRHydra
& 0.7248 & 0.4509 & 0.6327 & 0.7958 & 0.3646 \\

\bottomrule
\end{tabular}
\end{table*}

\begin{table*}[t]
\centering
\caption{Uncertainty and calibration performance on the New York City
extreme-weather dataset over the 2014--2021 test years. Results are averaged
across five training seeds for each year and then across years.}
\label{tab:nyc_uncertainty}
\begin{tabular}{lccc}
\toprule
Model & Brier & NLL & ECE \\
\midrule

ModernTCN
& 0.1238 & 0.3940 & 0.0776 \\

ModernTCN Forecast-then-Classify
& 0.1243 & 0.4009 & 0.0809 \\

ForeClassNet
& 0.1102 & 0.3507 & 0.0755 \\

Transformer Classifier
& 0.2359 & 1.2626 & 0.2226 \\

Transformer Forecast-then-Classify
& 0.2378 & 1.3102 & 0.2249 \\

ForeClassNet Dynamic
& \textbf{0.1093} & \textbf{0.3460} & 0.0705 \\

ForeClassNet MSE
& 0.1244 & 0.4021 & \textbf{0.0634} \\

ForeClassNet Energy
& 0.1292 & 0.4137 & 0.0702 \\

\bottomrule
\end{tabular}
\end{table*}

\begin{table*}[t]
\centering
\caption{Computational cost on the New York City extreme-weather dataset
over the 2014--2021 test years. Neural-network timings are averaged over
runs executed on the NVIDIA GeForce RTX 2080 Ti.
MRHydra timings are averaged across CPU runs.}
\label{tab:nyc_computational_cost}
\resizebox{\linewidth}{!}{%
\begin{tabular}{llrrr}
\toprule
Model & Hardware & Runs & Training (min) & Inference (ms/example) \\
\midrule

ModernTCN
& NVIDIA GeForce RTX 2080 Ti & 15 & 0.42 & 0.040 \\

ModernTCN Forecast-then-Classify
& NVIDIA GeForce RTX 2080 Ti & 5 & 0.44 & 0.047 \\

ForeClassNet
& NVIDIA GeForce RTX 2080 Ti & 7 & 1.11 & 3.821 \\

Transformer Classifier
& NVIDIA GeForce RTX 2080 Ti & 3 & 0.74 & 0.047 \\

Transformer Forecast-then-Classify
& NVIDIA GeForce RTX 2080 Ti & 10 & 1.03 & 0.054 \\

ForeClassNet Dynamic
& NVIDIA GeForce RTX 2080 Ti & 7 & 1.25 & 4.727 \\

ForeClassNet MSE
& NVIDIA GeForce RTX 2080 Ti & 7 & 5.42 & 3.768 \\

ForeClassNet Energy
& NVIDIA GeForce RTX 2080 Ti & 10 & 5.72 & 3.915 \\

MRHydra
& CPU & 40 & 3.27 & 2.291 \\

\bottomrule
\end{tabular}%
}
\end{table*}

Table~\ref{tab:nyc_performance} shows that the two primary ForeClassNet architectures achieve higher accuracy, F1 score, AUROC and average precision for both classes than the two ModernTCN models. ForeClassNet achieves an AUROC of 0.8921, class-0 AP of 0.9597, and class-1 AP of 0.7441, while ForeClassNet Dynamic achieves corresponding values of 0.8940, 0.9610, and 0.7499. The standard ForeClassNet achieves the highest accuracy of 0.8500 and F1 score of 0.6665, while ForeClassNet Dynamic achieves the highest AUROC and average precision for both classes among all evaluated models. The improvements in class-1 AP and AUROC of the primary ForeClassNet models compared to the ModernTCN models are statistically significant, with p-values of 0.0002 and 0.0008, respectively. Both primary architectures also outperform the Transformer and MRHydra models on these metrics. The ForeClassNet MSE and Energy variants achieve lower discrimination performance in this experiment. 

Table~\ref{tab:nyc_uncertainty} shows that both primary ForeClassNet architectures achieve lower Brier scores and NLL than the two ModernTCN baselines. ForeClassNet achieves a Brier score of 0.1102, NLL of 0.3507, and ECE of 0.0755, while ForeClassNet Dynamic achieves corresponding values of 0.1093, 0.3460, and 0.0705. ForeClassNet Dynamic achieves the lowest Brier score and NLL, while ForeClassNet MSE achieves the lowest ECE of 0.0634. The reductions in both Brier score and NLL for the primary ForeClassNet architectures against the ModernTCN models are statistically significant, with p-values of 0.0004 for each metric. The alternative ForeClassNet MSE and Energy variants achieve higher Brier scores and NLL than the two primary architectures, with ForeClassNet MSE achieving strong calibration. The two Transformer models have higher Brier scores, NLL, and ECE compared with the primary ForeClassNet architectures. 

\begin{table*}[t]
\centering
\caption{Classification performance on the Tokyo extreme-weather dataset
over the 2014--2021 test years. Results are averaged across five training seeds
for each year and then across years.}
\label{tab:tokyo_performance}
\begin{tabular}{lccccc}
\toprule
Model & Accuracy & F1 & AUROC & AP Class 0 & AP Class 1 \\
\midrule

ModernTCN
& 0.8546 & 0.6832 & 0.9171 & 0.9742 & 0.7516 \\

ModernTCN Forecast-then-Classify
& 0.8567 & 0.6820 & 0.9255 & 0.9770 & 0.7674 \\

ForeClassNet
& 0.8739 & 0.7203
& \textbf{0.9340} & \textbf{0.9798} & \textbf{0.7903} \\

Transformer Classifier
& 0.7726 & 0.5207 & 0.8091 & 0.9292 & 0.5511 \\

Transformer Forecast-then-Classify
& 0.7663 & 0.5177 & 0.7956 & 0.9231 & 0.5297 \\

ForeClassNet Dynamic
& \textbf{0.8749} & \textbf{0.7208}
& 0.9329 & 0.9788 & 0.7883 \\

ForeClassNet MSE
& 0.8511 & 0.6672 & 0.9071 & 0.9708 & 0.7202 \\

ForeClassNet Energy
& 0.8524 & 0.6664 & 0.9067 & 0.9705 & 0.7182 \\

MRHydra
& 0.7758 & 0.5231 & 0.6996 & 0.8503 & 0.3952 \\

\bottomrule
\end{tabular}
\end{table*}

\begin{table*}[t]
\centering
\caption{Uncertainty and calibration performance on the Tokyo
extreme-weather dataset over the 2014--2021 test years.
Results are averaged across five training seeds for
each year and then across years.}
\label{tab:tokyo_uncertainty}
\begin{tabular}{lccc}
\toprule
Model & Brier & NLL & ECE \\
\midrule

ModernTCN
& 0.1018 & 0.3201 & 0.0791 \\

ModernTCN Forecast-then-Classify
& 0.0973 & 0.3016 & 0.0738 \\

ForeClassNet
& \textbf{0.0883} & \textbf{0.2842} & 0.0786 \\

Transformer Classifier
& 0.1919 & 0.9450 & 0.1840 \\

Transformer Forecast-then-Classify
& 0.1983 & 0.9974 & 0.1905 \\

ForeClassNet Dynamic
& 0.0893 & 0.2875 & 0.0750 \\

ForeClassNet MSE
& 0.1042 & 0.3302 & 0.0667 \\

ForeClassNet Energy
& 0.1053 & 0.3338 & \textbf{0.0650} \\

\bottomrule
\end{tabular}
\end{table*}

\begin{table*}[t]
\centering
\caption{Computational cost on the Tokyo extreme-weather dataset
over the 2014--2021 test years. Neural-network timings are averaged over
runs executed on the NVIDIA GeForce RTX 2080 Ti.
MRHydra timings are averaged across CPU runs.}
\label{tab:tokyo_computational_cost}
\resizebox{\linewidth}{!}{%
\begin{tabular}{llrrr}
\toprule
Model & Hardware & Runs & Training (min) & Inference (ms/example) \\
\midrule

ModernTCN
& NVIDIA GeForce RTX 2080 Ti & 11 & 0.42 & 0.037 \\

ModernTCN Forecast-then-Classify
& NVIDIA GeForce RTX 2080 Ti & 6 & 0.66 & 0.065 \\

ForeClassNet
& NVIDIA GeForce RTX 2080 Ti & 6 & 1.25 & 3.780 \\

Transformer Classifier
& NVIDIA GeForce RTX 2080 Ti & 8 & 0.61 & 0.039 \\

Transformer Forecast-then-Classify
& NVIDIA GeForce RTX 2080 Ti & 8 & 1.06 & 0.062 \\

ForeClassNet Dynamic
& NVIDIA GeForce RTX 2080 Ti & 15 & 1.43 & 4.942 \\

ForeClassNet MSE
& NVIDIA GeForce RTX 2080 Ti & 10 & 5.59 & 3.670 \\

ForeClassNet Energy
& NVIDIA GeForce RTX 2080 Ti & 7 & 5.67 & 3.429 \\

MRHydra
& CPU  & 40 & 3.75 & 2.648 \\

\bottomrule
\end{tabular}%
}
\end{table*}

Table~\ref{tab:tokyo_performance} shows that the two primary ForeClassNet architectures achieve higher accuracy, F1 score, AUROC, and average precision for both classes than the ModernTCN models. ForeClassNet achieves an AUROC of 0.9340, class-0 AP of 0.9798, and class-1 AP of 0.7903, while ForeClassNet Dynamic achieves corresponding values of 0.9329, 0.9788, and 0.7883. ForeClassNet Dynamic achieves the highest accuracy of 0.8749 and F1 score of 0.7208, while the standard ForeClassNet achieves the highest AUROC of 0.9340 and average precision for both classes. The improvements in class-1 AP and AUROC of the primary ForeClassNet models compared to the ModernTCN models are statistically significant, with p-values of 0.0032 and 0.0020, respectively. Both primary ForeClassNet models also outperform the Transformer and MRHydra models. The ForeClassNet MSE and Energy variants achieve lower performance.

Table~\ref{tab:tokyo_uncertainty} shows that both primary ForeClassNet architectures achieve lower Brier scores and NLL than the two ModernTCN baselines. ForeClassNet achieves a Brier score of 0.0883, NLL of 0.2842, and ECE of 0.0786, while ForeClassNet Dynamic achieves corresponding values of 0.0893, 0.2875, and 0.0750. The standard ForeClassNet achieves the lowest Brier score and NLL, although both primary architectures have slightly higher ECE than the ModernTCN Forecast-then-Classify model. The reductions in both Brier score and NLL of the two primary ForeClassNet models compared to the ModernTCN models are statistically significant with p-values of 0.0002 and 0.0004, respectively. The alternative ForeClassNet MSE and Energy variants achieve higher Brier scores and NLL, although ForeClassNet Energy achieves the lowest ECE. Both ForeClassNet architectures outperform the Transformer baselines in terms of Brier score, NLL, and ECE.

The complete computational results comprise 62, 80, and 79 distinct model-hardware combinations for Shanghai, New York City, and Tokyo, respectively, over the 2014--2021 test years. We therefore report neural-network timings on the NVIDIA GeForce RTX 2080 Ti, which is used across all architectures and cities, providing a common hardware reference comparison. The two primary ForeClassNet architectures require more computation time than the ModernTCN models, but their training times remain approximately 1--1.5 minutes, with inference taking approximately 4--5 milliseconds per example despite using 100 Monte Carlo dropout forward passes.

Taken together, the experiments in Shanghai, New York City, and Tokyo provide consistent evidence that the two primary ForeClassNet architectures improve classification discrimination and probabilistic prediction relative to competing methods. 
 
\subsection{Future stock price movement}

Predicting future stock-price movements is challenging because financial returns exhibit a low signal-to-noise ratio and substantial temporal variation. In this experiment, each observation corresponds to a company earnings event. The input consists of 40 trading days of six market variables preceding the event trading day: opening price, highest price, lowest price, closing price, adjusted closing price, and trading volume. For pre-market earnings announcements, this is the announcement date; for post-market announcements, it is the next trading day. The objective is to use this observed history to identify economically useful directional signals over the subsequent five trading days and use these signals to construct a stock portfolio. The stock-price and earnings data were obtained from the US historical stock prices with earnings data dataset, available on Kaggle \cite{austin2021stock}.

We define three classes according to the five-day adjusted-price movement from the event trading day to five trading days later. An observation is assigned to the Down class if the price decreases by at least $5\%$, to the Up class if it increases by at least $5\%$, and to the Flat class otherwise. We use a temporally out-of-distribution test set comprising 21,511 earnings events occurring after those used for training, of which 4,591 belong to the Down class, 13,160 to the Flat class, and 3,760 to the Up class. We apply a common screen to extreme stock-price movements that may reflect corporate actions or data-adjustment artifacts. Specifically, an earnings event is excluded if its adjusted closing price triples or falls by at least two-thirds on any day of the five-day holding period. This rule excludes 16 of the 21,511 events, leaving 21,495
events for the portfolio experiment.

Our principal objective in this setting is to determine whether the predicted classes can be converted into economically useful portfolio decisions. We construct a confidence-filtered long--short strategy that takes long positions in confidently predicted Up stocks and short positions in confidently predicted Down stocks. Stocks predicted as Flat, or for which the classification confidence is insufficient, receive zero weight. For stock $i$ associated with earnings-event date $d$,
let $\hat{c}_{i,d}$ denote its predicted class and $p_{i,k}$ its predicted probability for class $k$. We define the classification margin as
\begin{equation*}
    m_{i,d} = p_{i,\hat{c}_{i,d}} -\max_{k\neq\hat{c}_{i,d}}p_{i,k}.
\end{equation*}
Thus, the margin is the difference between the
largest and second-largest predicted class probabilities.
For each event date $d$, define the eligible long and short sets
\begin{align*}
L_d &= \{i:\hat{c}_{i,d}=\mathrm{Up},\ m_{i,d}\geq 0.10\},\\
S_d &= \{i:\hat{c}_{i,d}=\mathrm{Down},\ m_{i,d}\geq 0.10\}.
\end{align*}
The initial event-cohort weights are
\begin{equation*}
w_{i,d} = \begin{cases}
\displaystyle \frac{0.5}{|L_d|},
    & i\in L_d,\\[6pt]
\displaystyle -\frac{0.5}{|S_d|},
    & i\in S_d,\\[6pt]
0, & \text{otherwise}.
\end{cases}
\end{equation*}
A long or short allocation is made only when the corresponding eligible set is nonempty. The two sides are selected independently, so the absence of qualifying stocks on one side does not prevent investment on the other side. Unused capital remains in cash and earns a zero return.

Positions are established at the market close of the event trading day and closed at the market close five trading days later, earning the five intervening daily returns. New portfolios are formed on subsequent eligible event dates, so their holding periods may overlap. Since each portfolio is held for five trading days, one-fifth of current portfolio capital is allocated to each newly formed cohort. After scaling each event-cohort allocation by the five-day holding period, we enforce an aggregate gross-exposure limit of 5\% per stock across all overlapping positions. If the absolute exposure to a stock exceeds this limit before daily returns are realized, its active positions are reduced proportionally to satisfy the cap. These adjustments contribute to transaction costs. During the holding period, positions are marked to market, allowing their weights to drift with realized stock returns without rebalancing their initial allocations. Specifically, the signed market value of each position evolves according to 
\begin{equation*}
    H_{i,d,t+1}^{-} = H_{i,d,t}^{+}(1+r_{i,t}),
\end{equation*}
where $H_{i,d,t}^{+}$ denotes the position value after trading on day $t$, $H_{i,d,t+1}^{-}$ denotes its value before trading on the following day, and $r_{i,t}$ is the intervening stock return. Positions are then liquidated at the end of their five-day holding period. Following the turnover-based transaction-cost approach of \cite{conlon2025asset}, we assume a transaction cost of c=20 basis points per dollar traded. Let $TC_{t}$, $R_{t}^{\text{gross}}$, and $R_{t}^{\text{net}}$ denote the transaction cost, gross portfolio return given by $\sum\limits_{i}W_{i,t}r_{i,t}$, and net return, on trading day $t$. 

\begin{equation*}
    \operatorname{TC}_{t}
    = c\sum_i |W_{i,t}^{+}-W_{i,t}^{-}|,
    \qquad
    R_t^{\mathrm{gross}}
    = \sum_i W_{i,t}^{+}r_{i,t},
    \qquad
    R_t^{\mathrm{net}}
    = (1+R_t^{\mathrm{gross}})
      (1-\operatorname{TC}_t)-1.
\end{equation*}
Here, $W_{i,t}^{-}$ and $W_{i,t}^{+}$ denote the aggregate portfolio weights of stock $i$ immediately before and after trading, respectively, normalized by the same pre-trade portfolio equity. These weights aggregate positions across overlapping cohorts and incorporate changes in their market values during the holding period. Transaction costs are incurred only when positions are established, adjusted through new cohort allocations, or liquidated, with simultaneous opposing trades in the same stock netted across cohorts. To measure portfolio performance and portfolio risk, we report several metrics. Let $T$ denote the number of out-of-sample trading days. We define the mean and sample standard deviation of daily net portfolio returns as 
\begin{equation*}
    \bar{R}^{\mathrm{net}}
    := \frac{1}{T}\sum_{t=1}^{T}R_t^{\mathrm{net}},
    \qquad \sigma_R:= \sqrt{\frac{1}{T-1}
        \sum_{t=1}^{T}
        (R_t^{\mathrm{net}}-\bar{R}^{\mathrm{net}})^2}.
\end{equation*}
Annualized return measures the compounded portfolio return
expressed on an annual basis:
\begin{equation*}
    \mathrm{AnnualReturn}
    := (
        \prod_{t=1}^{T}(1+R_t^{\mathrm{net}})
    )^{252/T}-1.
\end{equation*}
The Sharpe ratio measures average portfolio return relative
to its volatility, using a zero benchmark return:
\begin{equation*}
    \mathrm{Sharpe}
    := \sqrt{252}\,
    \frac{\bar{R}^{\mathrm{net}}}{\sigma_R}.
\end{equation*}
Average turnover measures the average absolute change in
aggregate portfolio weights between immediately before and after trading on each day:
\begin{equation*}
    \mathrm{AvgTurnover}
    := \frac{1}{T}
    \sum_{t=1}^{T}\sum_i
    |W_{i,t}^{+}-W_{i,t}^{-}|.
\end{equation*}
Maximum drawdown measures the largest percentage decline
in portfolio value from a preceding peak:
\begin{equation*}
    \mathrm{MaxDrawdown}
    := \min_{1\leq t\leq T}
    (\frac{V_t}{\max_{0\leq u\leq t}V_u}-1),
    \qquad V_t = \prod_{u=1}^{t}(1+R_u^{\mathrm{net}}),
    \quad V_0=1.
\end{equation*}
To assess portfolio risk, annualized volatility measures
the overall variability of daily net returns:
\begin{equation*}
    \mathrm{AnnualizedVolatility}
    := \sqrt{252}\,\sigma_R.
\end{equation*}
Annualized downside deviation measures the magnitude of
negative daily returns relative to a zero-return threshold:
\begin{equation*}
    \mathrm{DownsideDeviation}
    = \sqrt{
        \frac{252}{T}
        \sum_{t=1}^{T}
        \min(R_t^{\mathrm{net}},0)^2
    }.
\end{equation*}
The 95\% expected shortfall (ES95) measures the average
loss magnitude among the worst 5\% of daily returns:
\begin{equation*}
    \mathrm{ES95}
    = \max\left\{
        0,\,
        -\frac{1}{|\mathcal I|}
        \sum_{t\in\mathcal I}R_t^{\mathrm{net}}
    \right\},
    \qquad
    \mathcal I
    = \{t:R_t^{\mathrm{net}}\leq q_{0.05}\},
\end{equation*}
where $q_{0.05}$ is the empirical fifth percentile
of daily net returns.

\begin{table*}[t]
\centering
\caption{Portfolio performance results are averaged across five training seeds.}
\label{tab:stock_portfolio_performance}
\resizebox{\linewidth}{!}{%
\begin{tabular}{lrrrrr}
\toprule
Model
& Annual Return
& Sharpe
& Avg. Turnover
& Max Drawdown\\
\midrule
ModernTCN
& -0.01056 & -0.14626 & 0.01839
& -0.09603  \\

ModernTCN Forecast-then-Classify
& -0.00126 & -0.00992 & 0.01647
& -0.08114   \\

ForeClassNet
& \textbf{0.02478} & \textbf{0.64343} & 0.00932
& -0.04303  \\

Transformer Classifier
& -0.00111 & 0.30631 & 0.00147
& -0.02170  \\

Transformer Forecast-then-Classify
& 0.00789 & 0.53541 & 0.00140
& \textbf{-0.01255} \\

ForeClassNet Dynamic
& 0.00906 & 0.31288 & 0.00976
& -0.04313 \\

ForeClassNet MSE
& 0.00301 & 0.14648 & 0.01682
& -0.07256  \\

ForeClassNet Energy
& -0.01986 & -0.15482 & 0.01621
& -0.10234  \\

MultiRocket-Hydra
& -0.09241 & -0.88406 & 0.21247
& -0.30671  \\
\bottomrule
\end{tabular}%
}
\end{table*}

\begin{table*}[t]
\centering
\caption{Portfolio risk results are averaged across five training seeds.}
\label{tab:stock_portfolio_uncertainty}
\resizebox{\linewidth}{!}{%
\begin{tabular}{lrrr}
\toprule
Model
& Ann. Vol.
& Downside Dev.
& ES$_{95}$ \\
\midrule
ModernTCN
& 0.052342 & 0.037219 & 0.008711 \\

ModernTCN Forecast-then-Classify
& 0.051469 & 0.036497 & 0.008465 \\

ForeClassNet
& 0.036850 & 0.024833 & 0.005256 \\

Transformer Classifier
& 0.012930 & 0.008373 & 0.001330 \\

Transformer Forecast-then-Classify
& \textbf{0.011618} & \textbf{0.006436}
& \textbf{0.000344} \\

ForeClassNet Dynamic
& 0.037409 & 0.025285 & 0.005826 \\

ForeClassNet MSE
& 0.051186 & 0.035312 & 0.007799 \\

ForeClassNet Energy
& 0.050551 & 0.037304 & 0.008090 \\

MultiRocket-Hydra
& 0.103885 & 0.077883 & 0.015798 \\
\bottomrule
\end{tabular}%
}
\end{table*}

\begin{table*}[t]
\centering
\caption{Computational cost on the stock-price experiment. Times are averaged only across runs executed on the same recorded hardware.}
\label{tab:stock_computational_cost}

\resizebox{\linewidth}{!}{%
\begin{tabular}{llrrr}
\toprule
Model & Hardware & Runs & Training (min) & Inference (ms/example) \\
\midrule

ModernTCN
& NVIDIA GeForce GTX TITAN X
& 4
& 4.35
& 0.021 \\

ModernTCN
& NVIDIA TITAN Xp
& 1
& 3.08
& 0.024 \\

ModernTCN Forecast-then-Classify
& NVIDIA TITAN Xp
& 5
& 4.52
& 0.028 \\

ForeClassNet
& NVIDIA GeForce GTX 1080 Ti
& 2
& 10.14
& 4.385 \\

ForeClassNet
& NVIDIA TITAN Xp
& 2
& 8.66
& 3.400 \\

ForeClassNet
& Tesla T4
& 1
& 11.16
& 6.374 \\

Transformer Classifier
& NVIDIA GeForce GTX 1080 Ti
& 5
& 3.39
& 0.020 \\

Transformer Forecast-then-Classify
& NVIDIA GeForce GTX 1080 Ti
& 5
& 7.03
& 0.046 \\

ForeClassNet Dynamic
& NVIDIA GeForce RTX 2080 Ti
& 5
& 10.70
& 4.271 \\

ForeClassNet MSE
& NVIDIA GeForce RTX 2080 Ti
& 5
& 49.37
& 3.155 \\

ForeClassNet Energy
& NVIDIA GeForce RTX 2080 Ti
& 5
& 50.17
& 3.043 \\

MultiRocket-Hydra
& CPU
& 5
& 789.72
& 7.557 \\

\bottomrule
\end{tabular}%
}
\end{table*}

Table~\ref{tab:stock_portfolio_performance} reports the out-of-sample performance of the confidence-filtered long–short trading strategy. The standard ForeClassNet architecture achieves the highest average annualized return of 2.48\% and Sharpe ratio of 0.643 among the nine evaluated models. In comparison, ModernTCN achieves an annualized return of -1.06\% and a Sharpe ratio of -0.146, while ModernTCN Forecast-then-Classify achieves a Sharpe ratio of -0.010. ForeClassNet Dynamic also produces positive returns and a positive Sharpe ratio of 0.313. The improvements in Sharpe and annualized return of the primary ForeClassNet models compared to the ModernTCN models are statistically significant with p-values of 0.024 and 0.036 respectively.  

Table~\ref{tab:stock_portfolio_uncertainty} shows that ForeClassNet achieves an annualized volatility of 0.0369, downside deviation of 0.0248, and 95\% expected shortfall of 0.0053, while ForeClassNet Dynamic achieves corresponding values of 0.0374, 0.0253, and 0.0058. Both architectures outperform the two ModernTCN models on all three risk measures, with the strongest ModernTCN baseline achieving an annualized volatility of 0.0515, downside deviation of 0.0365, and expected shortfall of 0.0085. The reductions in annualized volatility, downside deviation, and expected shortfall of the primary ForeClassNet models compared to the ModernTCN models are statistically significant, with p-values of 0.0002, 0.0002, and 0.0002, respectively.

Table~\ref{tab:stock_computational_cost} shows that the two primary ForeClassNet architectures generally require more computation than the ModernTCN models. Across the recorded GPU configurations, ForeClassNet requires approximately 9--11 minutes of training and 3.4--6.4 milliseconds per example at inference, compared with approximately 3--5 minutes and 0.02--0.03 milliseconds for the ModernTCN baselines. The ForeClassNet MSE and Energy variants are considerably more expensive to train, requiring approximately 49--50 minutes, with inference times of 3.155 and 3.043 milliseconds, respectively. Nevertheless, the absolute computational costs of the primary ForeClassNet architectures remain relatively modest. 

Taken together, the stock-price experiment provides evidence that the primary ForeClassNet architectures can improve portfolio performance and reduce financial risk relative to the ModernTCN baselines.

\subsection{Sensitivity analysis}
In this section, we examine the impact of different hyperparameter choices for ForeClassNet on its classification performance. Specifically, we consider the following variants of ForeClassNet: 
\begin{itemize}
    \item [(a)] dropout probability of 0.05,
    \item [(b)] dropout probability of 0.2, 
    \item [(c)] Boltzmann temperature of 0.25, 
    \item [(d)] Boltzmann temperature of 0.5, 
    \item [(e)] Boltzmann temperature of 2, 
    \item [(f)] Boltzmann temperature of 4, 
    \item [(g)] filter sets $\{ 1, 3 \}, \{3,5 \}$, and $\{ 7,13 \}$,
    \item [(h)] filter sets $\{ 1,3,5 \}, \{ 1,3,5,7 \},\text{ and } \{ 5, 9, 13 \}$, 
    \item [(i)] filter sets $\{ 3,5,7 \}, \{ 3,5,9 \}, \text{ and } \{ 7, 13 , 19 \}$, 
    \item [(j)]  filter sets $\{ 1,3,5 \}, \{ 1,3,5,7 \}, \text{ and } \{3,7, 13,17  \}$, and
    \item [(k)] filter sets $\{ 3,5,7 \}, \{ 5,7,9 \}, \text{ and } \{ 11, 13, 21 \}$. 
\end{itemize}

The results are discussed below. 
\begin{table*}[t]
\centering
\caption{Sensitivity results for Oil Example 1. Results are
averaged over five training seeds.}
\label{tab:sensitivity_oil_1}
\small
\resizebox{\textwidth}{!}{%
\begin{tabular}{lrrrrrrr}
\toprule
Model & Accuracy & F1 & AUROC & AP$_1$ & Brier & NLL & ECE \\
\midrule
ForeClassNet & 0.8971 & 0.6207 & 0.9289 & 0.7543 & 0.0738 & 0.2388 & 0.0685 \\
\midrule
Dropout $p=0.05$ & 0.8953 & 0.6094 & 0.9214 & 0.7300 & 0.0750 & 0.2389 & 0.0581 \\
Dropout $p=0.20$ & \textbf{0.9131} & 0.6276 & 0.9228 & 0.7452 & \textbf{0.0686} & 0.2369 & 0.0633 \\
Temperature $T=0.25$ & 0.8973 & 0.6160 & 0.9281 & 0.7515 & 0.0726 & 0.2356 & 0.0629 \\
Temperature $T=0.50$ & 0.9048 & 0.6411 & 0.9304 & 0.7557 & 0.0699 & 0.2291 & 0.0631 \\
Temperature $T=2$ & 0.9037 & 0.6409 & 0.9300 & 0.7565 & 0.0704 & 0.2303 & 0.0647 \\
Temperature $T=4$ & 0.9055 & \textbf{0.6451} & 0.9317 & 0.7619 & 0.0696 & \textbf{0.2283} & 0.0638 \\
Filters G & 0.9037 & 0.6225 & 0.9231 & 0.7390 & 0.0708 & 0.2312 & \textbf{0.0559} \\
Filters H & 0.8855 & 0.6163 & 0.9261 & 0.7432 & 0.0791 & 0.2513 & 0.0836 \\
Filters I & 0.9058 & 0.6391 & 0.9301 & 0.7604 & 0.0697 & \textbf{0.2283} & 0.0611 \\
Filters J & 0.8960 & 0.6360 & 0.9306 & 0.7614 & 0.0739 & 0.2385 & 0.0741 \\
Filters K & 0.8962 & 0.6439 & 0.9318 & \textbf{0.7629} & 0.0739 & 0.2379 & 0.0790 \\
\bottomrule
\end{tabular}%
}
\end{table*}

\begin{table*}[t]
\centering
\caption{Sensitivity results for Oil Example 2. Results are
averaged over five training seeds.}
\label{tab:sensitivity_oil_2}
\small
\resizebox{\textwidth}{!}{%
\begin{tabular}{lrrrrrrrr}
\toprule
Model & Accuracy & F1 & AUROC & AP$_0$ & AP$_1$ & Brier & NLL & ECE \\
\midrule
ForeClassNet& 0.8665 & 0.8842 & 0.9562 & 0.9426 & 0.9692 & 0.0931 & 0.2944 & 0.0714 \\
\midrule
Dropout $p=0.05$ & \textbf{0.8802} & \textbf{0.8955} & \textbf{0.9605} & \textbf{0.9488} & \textbf{0.9720} & 0.0873 & 0.2745 & 0.0652 \\
Dropout $p=0.20$ & 0.8624 & 0.8752 & 0.9526 & 0.9373 & 0.9671 & 0.0905 & 0.2892 & 0.0538 \\
Temperature $T=0.25$ & 0.8693 & 0.8859 & 0.9570 & 0.9434 & 0.9699 & 0.0904 & 0.2839 & 0.0649 \\
Temperature $T=0.50$ & 0.8689 & 0.8861 & 0.9572 & 0.9438 & 0.9700 & 0.0913 & 0.2868 & 0.0679 \\
Temperature $T=2$ & 0.8702 & 0.8879 & 0.9574 & 0.9439 & 0.9701 & 0.0915 & 0.2868 & 0.0686 \\
Temperature $T=4$ & 0.8699 & 0.8868 & 0.9571 & 0.9436 & 0.9699 & 0.0915 & 0.2875 & 0.0666 \\
Filters G & 0.8710 & 0.8872 & 0.9576 & 0.9440 & 0.9703 & 0.0891 & 0.2794 & 0.0618 \\
Filters H & 0.8695 & 0.8856 & 0.9573 & 0.9438 & 0.9701 & 0.0896 & 0.2811 & 0.0628 \\
Filters I & 0.8707 & 0.8866 & 0.9579 & 0.9445 & 0.9705 & 0.0890 & 0.2805 & 0.0646 \\
Filters J & 0.8696 & 0.8862 & 0.9572 & 0.9437 & 0.9700 & 0.0906 & 0.2851 & 0.0639 \\
Filters K & 0.8703 & 0.8874 & 0.9573 & 0.9440 & 0.9701 & 0.0904 & 0.2831 & 0.0641 \\
\bottomrule
\end{tabular}%
}
\end{table*}

\begin{table*}[t]
\centering
\caption{Sensitivity results for Shanghai, 2014--2021.
Results are averaged over five training seeds and then
equally across test years.}
\label{tab:sensitivity_shanghai}
\small
\resizebox{\textwidth}{!}{%
\begin{tabular}{lrrrrrrrr}
\toprule
Model & Accuracy & F1 & AUROC & AP$_0$ & AP$_1$ & Brier & NLL & ECE \\
\midrule
ForeClassNet& 0.8512 & 0.5807 & 0.8865 & 0.9686 & 0.6793 & 0.1030 & 0.3283 & 0.0735 \\
\midrule
Dropout $p=0.05$ & 0.8484 & 0.5836 & 0.8861 & 0.9679 & 0.6749 & 0.1035 & 0.3263 & 0.0690 \\
Dropout $p=0.20$ & 0.8532 & 0.5696 & 0.8830 & 0.9665 & 0.6721 & 0.1055 & 0.3491 & 0.0932 \\
Temperature $T=0.25$ & 0.8521 & 0.5845 & 0.8874 & \textbf{0.9692} & 0.6785 & 0.1028 & 0.3283 & 0.0735 \\
Temperature $T=0.50$ & \textbf{0.8543} & 0.5928 & 0.8881 & 0.9688 & 0.6772 & 0.1026 & 0.3278 & 0.0728 \\
Temperature $T=2$ & 0.8533 & 0.5877 & \textbf{0.8899} & \textbf{0.9692} & 0.6775 & \textbf{0.1017} & \textbf{0.3251} & 0.0735 \\
Temperature $T=4$ & 0.8515 & 0.5807 & 0.8885 & 0.9687 & 0.6764 & 0.1023 & 0.3268 & 0.0724 \\
Filters G & 0.8513 & 0.5865 & 0.8883 & 0.9686 & 0.6778 & 0.1029 & 0.3285 & 0.0738 \\
Filters H & 0.8504 & 0.5860 & 0.8846 & 0.9679 & 0.6729 & 0.1041 & 0.3325 & 0.0757 \\
Filters I & 0.8523 & \textbf{0.5946} & 0.8889 & 0.9684 & \textbf{0.6846} & 0.1023 & 0.3276 & 0.0720 \\
Filters J & 0.8511 & 0.5877 & 0.8887 & 0.9686 & 0.6811 & 0.1024 & 0.3274 & 0.0735 \\
Filters K & 0.8525 & 0.5783 & 0.8854 & 0.9674 & 0.6696 & 0.1026 & 0.3277 & \textbf{0.0661} \\
\bottomrule
\end{tabular}%
}
\end{table*}

\begin{table*}[t]
\centering
\caption{Sensitivity results for NYC, 2014--2021.
Results are averaged over five training seeds and then
equally across test years.}
\label{tab:sensitivity_nyc}
\small
\resizebox{\textwidth}{!}{%
\begin{tabular}{lrrrrrrrr}
\toprule
Model & Accuracy & F1 & AUROC & AP$_0$ & AP$_1$ & Brier & NLL & ECE \\
\midrule
ForeClassNet & 0.8500 & 0.6665 & 0.8921 & 0.9597 & 0.7441 & 0.1102 & 0.3507 & 0.0755 \\
\midrule
Dropout $p=0.05$ & 0.8409 & 0.6488 & 0.8863 & 0.9584 & 0.7298 & 0.1127 & 0.3548 & \textbf{0.0682} \\
Dropout $p=0.20$ & \textbf{0.8545} & 0.6647 & 0.8911 & 0.9563 & 0.7486 & 0.1108 & 0.3631 & 0.0899 \\
Temperature $T=0.25$ & 0.8529 & 0.6720 & 0.8912 & 0.9592 & 0.7438 & 0.1094 & 0.3488 & 0.0730 \\
Temperature $T=0.50$ & 0.8500 & 0.6675 & 0.8918 & 0.9596 & 0.7439 & 0.1098 & 0.3496 & 0.0752 \\
Temperature $T=2$ & 0.8464 & 0.6625 & 0.8915 & 0.9596 & 0.7426 & 0.1104 & 0.3504 & 0.0731 \\
Temperature $T=4$ & 0.8526 & \textbf{0.6753} & 0.8928 & 0.9600 & 0.7474 & \textbf{0.1093} & 0.3482 & 0.0741 \\
Filters G & 0.8486 & 0.6700 & \textbf{0.8941} & 0.9603 & \textbf{0.7505} & 0.1098 & 0.3493 & 0.0748 \\
Filters H & 0.8474 & 0.6648 & 0.8911 & 0.9593 & 0.7441 & 0.1115 & 0.3531 & 0.0769 \\
Filters I & 0.8477 & 0.6707 & 0.8906 & 0.9592 & 0.7414 & 0.1105 & 0.3512 & 0.0743 \\
Filters J & 0.8510 & 0.6684 & 0.8913 & 0.9593 & 0.7444 & 0.1097 & 0.3488 & 0.0742 \\
Filters K & 0.8464 & 0.6625 & 0.8916 & 0.9587 & 0.7464 & 0.1104 & 0.3509 & 0.0757 \\
\bottomrule
\end{tabular}%
}
\end{table*}

\begin{table*}[t]
\centering
\caption{Sensitivity results for Tokyo, 2014--2021.
Results are averaged over five training seeds and then
equally across test years.}
\label{tab:sensitivity_tokyo}
\small
\resizebox{\textwidth}{!}{%
\begin{tabular}{lrrrrrrrr}
\toprule
Model & Accuracy & F1 & AUROC & AP$_0$ & AP$_1$ & Brier & NLL & ECE \\
\midrule
ForeClassNet & 0.8739 & 0.7203 & 0.9340 & 0.9798 & \textbf{0.7903} & 0.0883 & 0.2842 & 0.0786 \\
\midrule
Dropout $p=0.05$ & 0.8745 & 0.7212 & 0.9328 & 0.9789 & 0.7885 & 0.0882 & \textbf{0.2811} & \textbf{0.0695} \\
Dropout $p=0.20$ & 0.8737 & 0.7141 & 0.9336 & 0.9798 & 0.7843 & 0.0914 & 0.3050 & 0.0956 \\
Temperature $T=0.25$ & 0.8757 & 0.7253 & 0.9347 & \textbf{0.9801} & 0.7898 & 0.0883 & 0.2844 & 0.0785 \\
Temperature $T=0.50$ & 0.8750 & 0.7250 & 0.9340 & 0.9797 & 0.7891 & 0.0891 & 0.2871 & 0.0811 \\
Temperature $T=2$ & \textbf{0.8769} & \textbf{0.7289} & 0.9322 & 0.9791 & 0.7857 & 0.0894 & 0.2877 & 0.0789 \\
Temperature $T=4$ & \textbf{0.8769} & 0.7272 & 0.9347 & 0.9800 & 0.7901 & \textbf{0.0879} & 0.2830 & 0.0779 \\
Filters G & 0.8704 & 0.7131 & 0.9317 & 0.9790 & 0.7842 & 0.0909 & 0.2912 & 0.0783 \\
Filters H & 0.8761 & 0.7240 & 0.9341 & 0.9799 & 0.7882 & 0.0887 & 0.2858 & 0.0786 \\
Filters I & 0.8749 & 0.7181 & \textbf{0.9351} & \textbf{0.9801} & 0.7893 & 0.0887 & 0.2846 & 0.0790 \\
Filters J & 0.8757 & 0.7272 & 0.9336 & 0.9794 & 0.7893 & 0.0887 & 0.2862 & 0.0775 \\
Filters K & 0.8749 & 0.7197 & 0.9325 & 0.9792 & 0.7863 & 0.0888 & 0.2854 & 0.0751 \\
\bottomrule
\end{tabular}%
}
\end{table*}

Across the five experiments, ForeClassNet's performance is relatively insensitive to the hyperparameter choices considered. In particular, AUROC varies by at most 1.04 percentage points across the eleven sensitivity configurations, with differences of 0.79, 0.69, 0.78, and 0.34 percentage points for Oil Example 2, Shanghai, NYC, and Tokyo, respectively. Positive-class average precision varies by at most 3.29 percentage points, with differences of less than 2.1 percentage points in four of the five experiments. The Brier score also remains relatively stable, with maximum differences between configurations of 0.0105, 0.0042, 0.0038, 0.0034, and 0.0035 for Oil Example 1, Oil Example 2, Shanghai, NYC, and Tokyo, respectively. Similarly, NLL varies by at most 0.0240 across all experiments, although increasing the dropout probability to 0.2 leads to a deterioration in NLL of approximately 6.3\% and 7.3\% relative to the standard configuration for Shanghai and Tokyo, respectively. Overall, these results suggest that ForeClassNet's classification and probabilistic performance is generally robust to the hyperparameter choices examined, particularly in terms of AUROC and average precision, although some settings have a more noticeable effect on probabilistic calibration.

\subsection{Ablation study}

In this section, we explore the impact of the different architectural components of ForeClassNet on downstream decision-making. We then discuss the possible economic mechanisms behind the impacts of these different architectural components.

\begin{table*}[t]
\centering
\caption{Results are averaged over five training seeds.}
\label{tab:finance_consolidated_ablation}
\small
\resizebox{\textwidth}{!}{%
\begin{tabular}{lrrrrr}
\toprule
Metric
    & Full Model
    & MC Off
    & Concrete Dropout
    & No Boltzmann
    & No Forecasting \\
\midrule

Annualized return
    & \textbf{0.02478}
    & 0.02030
    & 0.01057
    & 0.00453
    & -0.00782 \\

Sharpe ratio
    & \textbf{0.64343}
    & 0.52480
    & 0.38160
    & 0.17382
    & -0.16144 \\

Average turnover
    & 0.00932
    & 0.00889
    & \textbf{0.00710}
    & 0.01061
    & 0.01132 \\

Maximum drawdown
    & -0.04303
    & \textbf{-0.04282}
    & -0.04327
    & -0.05543
    & -0.06734 \\

\midrule

Annualized volatility
    & 0.03685
    & 0.03617
    & \textbf{0.03383}
    & 0.04025
    & 0.04419 \\

Downside deviation
    & 0.02483
    & 0.02476
    & \textbf{0.02231}
    & 0.02933
    & 0.03357 \\

Expected shortfall (ES95)
    & 0.00526
    & 0.00522
    & \textbf{0.00494}
    & 0.00639
    & 0.00767 \\

\bottomrule
\end{tabular}%
}
\end{table*}

Monte Carlo dropout improves the Sharpe ratio from 0.525 to 0.643. One possible explanation relates to the increased volatility inherent in financial time series. To investigate whether the benefit of MC dropout varies with individual-stock volatility, we calculate the historical return volatility over the 40 trading days preceding each earnings announcement and regress the paired NLL advantage of MC dropout over deterministic inference on this measure. We use winsorized log volatility, standardized to have unit standard deviation, and control for market volatility, realized earnings-outcome class, and calendar-year effects, with standard errors clustered by earnings date. The NLL advantage of MC dropout increases significantly with individual-stock volatility, with a coefficient of $0.00269$ ($p=0.0017$). This suggests that MC dropout may be particularly beneficial for probabilistic prediction in more volatile financial time series.

We then examine the effect of the dropout formulation on ForeClassNet by comparing the standard Monte Carlo dropout procedure, which uses a fixed dropout probability of 0.1, with Concrete dropout \cite{gal2017concrete}, which learns dropout probabilities during training. For Concrete dropout, we set the prior length scale to $l=0.01$ and the model precision to $\tau=1$, using fixed regularization settings across experiments. Both procedures average class probabilities over 100 stochastic forward passes at inference.

The ForeClassNet model with Monte Carlo dropout achieves a Sharpe ratio of 0.643, compared with 0.382 for Concrete dropout. One possible explanation relates to the nonstationarity of financial markets and the regularization learned by Concrete dropout. Specifically, Concrete dropout jointly optimizes the network parameters and layer-specific dropout probabilities over the training distribution. If the relationships between historical price movements and future returns differ across market regimes, the training objective may favor stronger regularization that suppresses patterns which are inconsistent across those regimes, even when such patterns contain useful predictive information during periods of heightened volatility. This is particularly relevant because Concrete dropout learns an average dropout probability of approximately 0.40 in its classification network, compared with the fixed probability of 0.1 used by standard Monte Carlo dropout. Neither method adapts its dropout probabilities to market conditions at inference, but the stronger regularization learned by Concrete dropout may make its predictions less sensitive to signals that become informative when market conditions change.

To investigate this possibility, we construct a time-varying measure of market volatility by calculating the cross-sectional median daily stock return and its rolling standard deviation over the preceding 20 trading days. This measure is updated over time and lagged by one trading day, ensuring that only information available before each earnings announcement is used. We then regress the paired NLL advantage of standard Monte Carlo dropout over Concrete dropout on this measure, controlling for individual-stock volatility over the preceding 40 trading days, realized earnings-outcome class, and calendar-year effects, with standard errors clustered by earnings date. Using standardized, winsorized log volatility measures, the NLL advantage of Monte Carlo dropout increases significantly with market volatility, with a coefficient of $0.00300$ ($p=0.000094$). Furthermore, the median historical stock volatility increases from $0.0185$ in the training period to $0.0228$ in the test period, suggesting a shift toward more volatile conditions. These findings are consistent with the possibility that the regularization learned by Concrete dropout is less effective when financial relationships change across market regimes.

Finally, we examine the impact of the Boltzmann convolutions and forecasting branch. The full ForeClassNet model achieves an average Sharpe ratio of 0.64343, compared with 0.17382 when the Boltzmann convolutions are removed and -0.16144 when the forecasting branch is removed. The corresponding annualized returns are 2.478\%, 0.453\%, and -0.782\%, respectively.

One possible explanation for these differences relates to how information is incorporated into stock prices and trading activity before earnings announcements. Financial information is not necessarily reflected in prices smoothly or at a single temporal scale. Some information is incorporated through overnight price adjustments, while other signals emerge through the interaction between price movements and trading volume during regular trading. The Boltzmann convolutions combine information across multiple temporal scales, potentially allowing ForeClassNet to better represent these different patterns of information incorporation. Similarly, the forecasting branch encourages the model to learn the temporal evolution of the observed financial variables, which may provide useful additional supervision when anticipating price movements surrounding earnings announcements.

To investigate these possibilities, we construct two measures using the 40 trading days preceding each earnings announcement. The first measures the proportion of absolute overnight and intraday price movements attributable to overnight adjustments, while the second measures the correlation between daily stock returns and log trading volume. We then examine whether the NLL advantage of the full ForeClassNet model over each architectural ablation varies with these characteristics. We estimate regressions including both measures, controlling for historical stock volatility, downside return variation, recent price drift, and calendar-year effects, with standard errors clustered by earnings date. For the forecasting branch, the NLL advantage increases significantly with the overnight information share among stocks experiencing upward earnings-related price movements, with a coefficient of $0.00230$ ($p=0.0133$). A stronger positive return--volume relationship is also associated with a significantly greater forecasting advantage for these observations, with a coefficient of $0.00179$ ($p=0.0079$).

For the Boltzmann convolutions, the NLL advantage likewise increases with the strength of the return--volume relationship among upward earnings outcomes, with a coefficient of $0.00221$ ($p=0.0050$). This association is also statistically significant when the test period is divided into two temporal halves, with $p=0.0311$ and $p=0.0323$, respectively. We additionally find evidence of a positive relationship between the Boltzmann advantage and overnight information incorporation across all earnings outcomes, with a coefficient of $0.00055$ ($p=0.0506$). Taken together, these findings are consistent with the possibility that the forecasting branch and Boltzmann convolutions improve probabilistic predictions by better representing financial signals associated with overnight information incorporation and the interaction between price movements and trading activity. Such signals may be particularly relevant when anticipating earnings-related price movements, providing a possible economic explanation for the improved downstream portfolio performance of the full ForeClassNet architecture.

\section{Conclusion}
We introduced Foreclassing, a classification problem in which labels depend on future time series observations that are unavailable at prediction time. We provided a mathematical formulation and established theoretical results characterizing Bayes classification risk, classification error bounds, and potential benefits of using realized future observations during training. We also proposed ForeClassNet, a family of neural networks that jointly perform forecasting and classification while accounting for predictive uncertainty. Experiments involving power transformers, extreme weather, and financial markets demonstrated the potential of this approach to improve classification performance, probabilistic prediction, and downstream decision-making. Future research could extend the theoretical results by deriving further finite-sample error bounds. Another direction for future research is to consider Foreclassing for recursive multi-step forecasting and how forecasting horizon length and error accumulation affect downstream classification performance. Another direction is to investigate learning objectives that explicitly account for the potential costs and opportunity costs of downstream decision-making, rather than focusing on forecasting and classification accuracy. 

\bibliographystyle{plainnat}
\bibliography{bibliography}
\section{Appendix}
\subsection{Proof of Theorem 1}
\begin{proof}
Let $\eta^{(h)} = \mathbb{E}[Y|\mathcal{F}_{t}^{(h)}]$ which means $\eta^{(h)}$ is $\mathcal{F}_{t}^{(h)}$ measurable. Furthermore, $\phi(p)$ is Borel-measurable. Therefore, $\phi(\eta^{(h)})$ is $\mathcal{F}_{t}^{(h)}$ measurable. Additionally, $\mathbb{E}[|\phi(\eta^{(h)})|] \leq \frac{1}{2} < \infty$ because $0 \leq \phi \leq \frac{1}{2}$. We know $\eta^{(h)}$ is measurable and $\mathbb{E}[|\eta^{(h)}|] \leq 1 < \infty$ since $\eta^{(h)}$ is $[0,1]$-valued. By the law of total expectation, 
\begin{equation*}
    \mathbb{E}[\eta^{(h+1)}|\mathcal{F}_{t}^{(h)}] = \mathbb{E}[\mathbb{E}[Y|\mathcal{F}_{t}^{(h+1)}]|\mathcal{F}_{t}^{(h)}] = \mathbb{E}[Y|\mathcal{F}_{t}^{(h)}] = \eta^{(h)} \qquad a.s.
\end{equation*}
Therefore, \((\eta^{(h)})_{h\ge0}\) is a martingale with respect to \((\mathcal F_t^{(h)})_{h\ge0}\). Note that $\phi(p)$ is concave. Then, 
\begin{equation*}
     \mathbb{E}[\phi(\eta^{(h+1)})|\mathcal{F}_{t}^{{(h)}}] \leq \phi(\mathbb{E}[\eta^{(h+1)}|\mathcal{F}_{t}^{(h)}]) = \phi(\eta^{(h)}),
\end{equation*}
where the inequality is by Jensen's inequality for concave functions and the last equality is because $\eta^{(h)}$ is a martingale. Then, by definition, $\phi(\eta^{(h)})$ is a supermartingale. This proves part (a) of the statement. Now, 
\begin{align*}
    R_{0-1}(\mathcal{F}_{t}^{(h)}) &= \mathbb{E}[\phi(\eta^{(h)})] \\
    \mathbb{E}[\phi(\eta^{(h+1)})|\mathcal{F}_{t}^{(h)}] & \leq \phi(\eta^{(h)}), \text{ since } (\phi(\eta^{(h)}))_{h \geq 0} \text{ is a supermartingale. }\\
    \text{This implies } \mathbb{E}[\mathbb{E}[\phi(\eta^{(h+1)})|\mathcal{F}_{t}^{(h)}]] &\leq \mathbb{E}[\phi(\eta^{(h)})],\text{ which implies }\\
    R_{0-1}(\mathcal{F}_{t}^{(h+1)})=\mathbb{E}[\phi(\eta^{(h+1)})] &\leq \mathbb{E}[\phi(\eta^{(h)})]   = R_{0-1}(\mathcal{F}_{t}^{(h)}),\text{ which proves part b}.
\end{align*}
Let $\Delta_{h} := \phi(\eta^{(h)}) - \mathbb{E}[\phi(\eta^{(h+1)})|\mathcal{F}_{t}^{(h)}] \geq 0 $ a.s.\\
$R_{0-1}(\mathcal{F}_{t}^{(h)}) = R_{0-1}(\mathcal{F}_{t}^{(h+1)}) \iff \mathbb{E}[\phi(\eta^{(h)})] = \mathbb{E}[\phi(\eta^{(h+1)})] \iff \mathbb{E}[\Delta_{h}] = 0 \iff \Delta_{h} = 0$ a.s. 
\begin{align*}
    \Delta_{h} &= \phi(\eta^{(h)}) - \mathbb{E}[\phi(\eta^{(h+1)})|\mathcal{F}_{t}^{(h)}]\\ &  = (\frac{1}{2} - |\eta^{(h)} - \frac{1}{2}|) - \mathbb{E}[\frac{1}{2} - |\eta^{(h+1)} - \frac{1}{2}||\mathcal{F}_{t}^{(h)}] \\ &= \mathbb{E}[|\eta^{(h+1)} - \frac{1}{2}||\mathcal{F}_{t}^{(h)}] - |\eta^{(h)} - \frac{1}{2}|
\end{align*}

Since $\eta^{(h)}$ is a martingale, we know $\mathbb{E}[\eta^{(h+1)} - \frac{1}{2}|\mathcal{F}_{t}^{(h)}]  = \eta^{(h)} - \frac{1}{2}$. 

$\mathbb{E}[|\eta^{(h+1)}- \frac{1}{2}||\mathcal{F}_{t}^{(h)}] = |\mathbb{E}[\eta^{(h+1)} - \frac{1}{2}|\mathcal{F}_{t}^{(h)}]| \iff 
\begin{cases}
\eta^{(h+1)} - \frac{1}{2} \geq 0 & \text{a.s. on } \{ \mathbb{E}[\eta^{(h+1)}-\frac{1}{2}|\mathcal{F}_{t}^{(h)}]>0 \},\\
\eta^{(h+1)} - \frac{1}{2} \leq 0, & \text{a.s. on } \{ \mathbb{E}[\eta^{(h+1)}-\frac{1}{2}|\mathcal{F}_{t}^{(h)}]<0 \},\\
\eta^{(h+1)} - \frac{1}{2} = 0, & \text{a.s. on  } \{ \mathbb{E}[\eta^{(h+1)}-\frac{1}{2}|\mathcal{F}_{t}^{(h)}]=0 \}.
\end{cases}$

Then, $R_{0-1}(\mathcal{F}_{t}^{(h+1)}) = R_{0-1}(\mathcal{F}_{t}^{(h)}) \iff \begin{cases}
    \eta^{(h)}> \frac{1}{2} \implies \eta^{(h+1)} \geq \frac{1}{2} \; a.s. \\
    \eta^{(h)}< \frac{1}{2} \implies \eta^{(h+1)} \leq \frac{1}{2} \; a.s \\
    \eta^{(h)} = \frac{1}{2} \implies \eta^{(h+1)} = \frac{1}{2} \; a.s.
\end{cases}$.
Equivalently, 
\begin{equation*}
    R_{0-1}(\mathcal{F}_{t}^{(h+1)}) = R_{0-1}(\mathcal{F}_{t}^{(h)})
\end{equation*}
if and only if 
\begin{equation*}
    \mathbb{P}(\{ \eta^{(h)}>\frac{1}{2} , \eta^{(h+1)} < \frac{1}{2} \} \cup \{\eta^{(h)}< \frac{1}{2}, \eta^{(h+1)} > \frac{1}{2} \} \cup \{ \eta^{(h)} = \frac{1}{2} , \eta^{(h+1)} \neq \frac{1}{2} \}) =0.
\end{equation*}
By part (b), the Bayes risk cannot increase when the information set is enlarged. Consequently, strict inequality occurs precisely when equality fails. From the equality characterization established above, this gives 
\begin{equation*}
    R_{0-1}(\mathcal{F}_{t}^{(h+1)}) < R_{0-1}(\mathcal{F}_{t}^{(h)}) \iff R_{0-1}(\mathcal{F}_{t}^{(h+1)}) \neq R_{0-1}(\mathcal{F}_{t}^{(h)}) \iff \mathbb{P}(A_{h})>0. 
\end{equation*}
This proves part (c).
\end{proof}
\subsection{Proof of Result 1}
By $((X_{i}, Z_{i}, Y_{i}))_{i=1}^{n} \perp\!\!\!\perp (X,Z,Y)$ and the training and test examples are independent of $U$, we have that 
\begin{equation*}
    \mathbb{E}[Y|X,D,U] = \mathbb{E}[Y|X] = \eta_{0}(X).
\end{equation*}
Let $T_{D}(\cdot,\cdot)$ denote a forecasting function fitted using training data $D$. Clearly $\sigma(X,D, \tilde{Z}) \subseteq \sigma(X, D, U)$. Thus, by the law of total expectation, 
\begin{equation*}
    \mathbb{E}[Y|X,D, \tilde{Z}] = \mathbb{E}[\mathbb{E}[Y|X,D,U]|X, D, \tilde{Z}] = \mathbb{E}[\eta_{0}(X)|X, D, \tilde{Z}] = \eta_{0}(X). 
\end{equation*}
Similarly, 
\begin{equation*}
    \mathbb{E}[\eta_{1}(X,Z)|X] = \mathbb{E}[\mathbb{E}[Y|X,Z]|X]  = \eta_{0}(X). 
\end{equation*}
By Jensen's inequality and the concavity of $\phi$, we have 
\begin{equation*}
    \mathbb{E}[\phi(\eta_{1}( X,Z))|X] \leq \phi(\mathbb{E}[\eta_{1}(X,Z)|X]) = \phi(\eta_{0}(X)). 
\end{equation*}
Thus, 
\begin{equation*}
    R_{XZ}^{*} \leq R_{X}^{*}. 
\end{equation*}
For $a \in \{0,1 \}$ and $p \in [0,1]$ define 
\begin{equation*}
    r(a;p) = p(1-a) + (1-p)a. 
\end{equation*}
Then 
\begin{equation*}
    r(a;p) - \phi(p)=
\begin{cases}
(2p-1) \boldsymbol{1}\{ a = 0 \}, & \text{if } p>\frac{1}{2},\\
(1-2p)\boldsymbol{1}\{ a=1 \}, & \text{if } p< \frac{1}{2},\\
0, & \text{if } p = \frac{1}{2}.
\end{cases}
\end{equation*}
Therefore, 
\begin{equation*}
    r(a;p) - \phi(p) = |2p-1| \boldsymbol{1}\{ a \neq \boldsymbol{1}\{ p \geq \frac{1}{2} \} \}.
\end{equation*}
Then conditional on $(X,D,U)$, the error probability for a classification rule trained on data $D$ with stochastic prediction $U$, $g_{D}(X,U)$, is given by $r(g_{D}(X,U); \eta_{0}(X))$. Then
\begin{align*}
    R_{D}(g_{D}) - R_{X}^{*} &= \mathbb{E}_{X,Y, U}[\boldsymbol{1} \{ g_{D}(X,U) \neq Y \}|D] - \mathbb{E}_{X}[\phi(\eta_{0}(X))]. \\
    &= \mathbb{E}_{X,U}[\mathbb{E}[\boldsymbol{1}\{g_{D}(X,U) \neq Y \}|X,D,U] - \phi(\eta_{0}(X))|D]\\
    &=\mathbb{E}_{X,U}[r(g_{D}(X,U); \eta_{0}(X)) - \phi(\eta_{0}(X))|D]\\
    & = \mathbb{E}_{X,U}[(2\eta_{0}(X)-1) \boldsymbol{1}\{ \eta_{0}(X) > \frac{1}{2} , g_{D}(X,U) = 0\}\\
    & + (1-2\eta_{0}(X))\boldsymbol{1} \{ \eta_{0}(X) < \frac{1}{2}, g_{D}(X,U) = 1 \}|D]\\
    & = \mathbb{E}_{X,U}[|2\eta_{0}(X)-1| (\boldsymbol{1}\{ \eta_{0}(X) > \frac{1}{2}, g_{D}(X,U) = 0 \} + \boldsymbol{1}\{ \eta_{0}(X) < \frac{1}{2}, g_{D}(X,U) = 1 \})|D]\\
    &= \mathbb{E}_{X,U}[|2\eta_{0}(X)-1| (\boldsymbol{1} \{ \eta_{0}(X) \geq \frac{1}{2}, g_{D}(X,U) =0 \} + \boldsymbol{1} \{ \eta_{0}(X) < \frac{1}{2}, g_{D}(X,U) = 1 \} ) |D].
\end{align*}
Since the events are disjoint 
\begin{align*}
    \boldsymbol{1} \{ \eta_{0}(X) \geq \frac{1}{2}, g_{D}(X,U) = 0 \} + \boldsymbol{1}\{ \eta_{0}(X)  < \frac{1}{2}, g_{D}(X,U) = 1\}  &= \boldsymbol{1} \{ \{ \eta_{0}(X) \geq \frac{1}{2}, g_{D}(X,U) = 0 \}\\ &\cup \{ \eta_{0}(X) < \frac{1}{2}, g_{D}(X,U) =1 \} \}\\
    &= \boldsymbol{1}\{ g_{D}(X,U) \neq \boldsymbol{1}\{ \eta_{0}(X) \geq \frac{1}{2} \} \}. 
\end{align*}
Thus, 
\begin{equation*}
    R_{D}(g_{D}) - R_{X}^{*} = \mathbb{E}_{X,U}[|2\eta_{0}(X) - 1|\boldsymbol{1}\{ g_{D}(X,U) \neq \boldsymbol{1}\{ \eta_{0}(X) \geq \frac{1}{2} \} \}|D]. 
\end{equation*}
Then by adding $R_{X}^{*} - R_{XZ}^{*}$ to both sides we obtain 
\begin{equation*}
    R_{D}(g_{D}) - R_{XZ}^{*} = R_{X}^{*} - R_{XZ}^{*} + \mathbb{E}_{X,U}[|2\eta_{0}(X) - 1| \boldsymbol{1}\{g_{D}(X,U) \neq \boldsymbol{1}\{\eta_{0}(X) \geq \frac{1}{2} \} \}|D].
\end{equation*}
That is 
\begin{equation*}
    R_{D}(g_{D}) - R_{XZ}^{*} = (R_{X}^{*} - R_{XZ}^{*}) +  (R_{D}(g_{D}) - R_{X}^{*}). 
\end{equation*}
\subsection*{Proof of Result 2}
Consider $v,p \in [0,1]$ and define 
\begin{equation*}
    a(v) := \boldsymbol{1}\{ v \geq \frac{1}{2} \} \qquad b(p):= \boldsymbol{1}\{ p \geq \frac{1}{2} \}.
\end{equation*}
Then, 
\begin{equation*}
    |2p-1| \boldsymbol{1}\{ a(v) \neq b(p)\} \leq 2|v-p|. 
\end{equation*}
First if $a(v) = b(p)$, the left-hand side is zero, so the inequality holds. Second, if $a(v) \neq b(p)$ and $ p > 1/2$, then $b(p) =1$ and $a(v) = 0$. This implies $v < 1/2$. Therefore, 
\begin{equation*}
    |v-p| = |p-v| = (p-\frac{1}{2}) + (\frac{1}{2} - v) \geq p - \frac{1}{2}.
\end{equation*}
Then multiplying by 2 yields the inequality. Thirdly, we can use the same argument for $a(v) \neq b(p)$ and $p < 1/2 $ which also yields the inequality. Finally, if $p = 1/2$, then $|2p-1|=0$, so the inequality holds regardless of whether $a(v)=b(p)$. Now consider the fitted function $V_{D}$ trained on dataset $D$ which outputs class probabilities. Then define 
\begin{equation*}
    g_{D}^{V}(x,u) = \begin{cases}
0, V_{D}(x,u) < \frac{1}{2},\\
1, V_{D}(x,u)  \geq \frac{1}{2}.
\end{cases}
\end{equation*}
Define 
\begin{equation*}
    g_{X}^{*}(x):= \boldsymbol{1}\{ \eta_{0}(x) \geq \frac{1}{2} \}.
\end{equation*}
By Result 1, 
\begin{equation*}
    R_{D}(g_{D}^{V}) - R_{X}^{*} = \mathbb{E}_{X,U}[|2\eta_{0}(X) -1| \boldsymbol{1}\{ g_{D}^{V}(X,U) \neq g_{X}^{*}(X) \} |D]. 
\end{equation*}
Note by the inequality above 
\begin{equation*}
    |2\eta_{0}(X)-1| \boldsymbol{1}\{ g_{D}^{V}(X,U) \neq g_{X}^{*}(X) \} \leq 2 |V_{D}(X,U) - \eta_{0}(X)|.
\end{equation*}
Then taking conditional expectations gives
\begin{equation*}
    0 \leq R_{D}(g_{D}^{V}) - R_{X}^{*} \leq 2\mathbb{E}_{X,U} [|V_{D}(X,U) - \eta_{0}(X)|D]. 
\end{equation*}
Fix the training dataset $D$ and history value $x$. By the tower property and the definitions of $s_{D}$ and $\eta_{1}$, we obtain
\begin{align*}
    \mathbb{E}_{U}[V_{D}(x,U)] - \eta_{0}(x) &= \mathbb{E}_{U}[V_{D}(x,U)] - \mathbb{E}[Y|X=x]\\
    &= \mathbb{E}[s_{D}(x,\tilde{Z})| X=x , D]- \mathbb{E}[\eta_{1}(x,Z)|X=x]\\
    & = \int\limits_{\mathcal{Z}} s_{D}(x,z)Q_{D,x}(dz) - \int\limits_{\mathcal{Z}}\eta_{1}(x,z) P_{x}(dz). 
\end{align*}
where $Q_{D,x}(dz)$ and $P_{x}(dz)$ denote the conditional probability measures for the forecasted future $\tilde{z}$ given $D$ and $x$, and the actual future $Z$ given $x$, respectively. Then 
\begin{align*}
    \mathbb{E}_{U}[V_{D}(x,U)] - \eta_{0}(x) &= \int\limits_{\mathcal{Z}} s_{D}(x,z) Q_{D,x}(dz) - \int\limits_{\mathcal{Z}}s_{D}(x,z)P_{x}(dz) + \int\limits_{\mathcal{Z}}s_{D}(x,z) P_{x}(dz)\\& - \int\limits_{\mathcal{Z}}\eta_{1}(x,z)P_{x}(dz)\\
    &=\int\limits_{\mathcal{Z}} s_{D}(x,z) (Q_{D,x} - P_{x})(dz) + \int\limits_{\mathcal{Z}}[s_{D}(x,z) - \eta_{1}(x,z)]P_{x}(dz). 
\end{align*}
By the triangle inequality, 
\begin{align*}
    |\mathbb{E}_{U}[V_{D}(x,U)] - \eta_{0}(x)| &= |\int\limits_{\mathcal{Z}}s_{D}(x,z)(Q_{D,x} - P_{x})(dz) + \int\limits_{\mathcal{Z}}[s_{D}(x,z) - \eta_{1}(x,z)] P_{x}(dz)|\\
    & \leq |\int\limits_{\mathcal{Z}} s_{D}(x,z) (Q_{D,x} - P_{x})(dz)| + |\int\limits_{\mathcal{Z}} [s_{D}(x,z) - \eta_{1}(x,z)]P_{x}(dz)|\\
    &\leq |\int\limits_{\mathcal{Z}}s_{D}(x,z)(Q_{D,x} - P_{x})(dz)| + \int\limits_{\mathcal{Z}}|s_{D}(x,z) - \eta_{1}(x,z)|P_{x}(dz)\\
    &= |\int\limits_{\mathcal{Z}} \int\limits_{0}^{1} \boldsymbol{1} \{ s_{D}(x,z) >r \}dr (Q_{D,x} - P_{x})(dz)| + \int \limits_{\mathcal{Z}} |s_{D}(x,z) - \eta_{1}(x,z)|P_{x}(dz).  
\end{align*}
By Tonelli, 
\begin{align*}
    \int\limits_{\mathcal{Z}} \int\limits_{0}^{1} \boldsymbol{1}\{ s_{D}(x,z) > r \} dr Q_{D,x}(dz) &= \int\limits_{0}^{1} \int\limits_{\mathcal{Z}} \boldsymbol{1} \{ s_{D}(x,z) > r \} Q_{D,x}(dz) dr \\
    &= \int\limits_{0}^{1} Q_{D,x} (\{ z: s_{D}(x,z)>r \}) dr
\end{align*}
and\begin{equation*}
    \int\limits_{\mathcal{Z}}s_{D}(x,z) P_{x}(dz) = \int\limits_{0}^{1}P_{x}(\{z: s_{D}(x,z) >r \})dr.
\end{equation*}
Then 
\begin{align*}
    \int\limits_{\mathcal{Z}} s_{D}(x,z) (Q_{D,x}-P_{x})(dz) &= \int\limits_{0}^{1} Q_{D,x}(\{ z: s_{D}(x,z)>r \})dr - \int\limits_{0}^{1} P_{x}(\{ z: s_{D}(x,z) >r \}) dr\\
    &= \int\limits_{0}^{1} (Q_{D,x} - P_{x}) (\{ z: s_{D}(x,z) >r \})dr.
\end{align*}
We then have 
\begin{align*}
    |\int\limits_{\mathcal{Z}} s_{D}(x,z) (Q_{D,x}-P_{x})(dz)| & \leq \int\limits_{0}^{1} |(Q_{D,x} - P_{x})(\{z: s_{D}(x,z)>r \})|dr\\
    & \leq \int\limits_{0}^{1} TV(P_{x}, Q_{D,x})dr\\
    &= TV(P_{x}, Q_{D,x}) \int\limits_{0}^{1}1 dr\\
    &= TV(P_{x}, Q_{D,x}). 
\end{align*}
Hence
\begin{equation*}
    |\mathbb{E}_{U}[V_{D}(x,U)] - \eta_{0}(x)| \leq TV(P_{x}, Q_{D,x}) + \int\limits_{\mathcal{Z}} |s_{D}(x,z) - \eta_{1} (x,z)|P_{x}(dz). 
\end{equation*}
Then 
\begin{equation*}
    \mathbb{E}_{X}[|\mathbb{E}_{U}[V_{D}(X,U)] - \eta_{0}(X)||D] \leq \int\limits_{\mathcal{X}} TV(P_{x}, Q_{D,x})\mu(dx) + \int\limits_{\mathcal{X}} \int\limits_{\mathcal{Z}} |s_{D}(x,z) - \eta_{1}(x,z)| P_{x}(dz)\mu(dx). 
\end{equation*}
Let $U_{1}, \dots, U_{M}$ be the independent draws with the same distribution as $U$. Then, for fixed $x$ and $D$
\begin{align*}
    \mathbb{E}_{U_{1}, \dots, U_{M}}[\frac{1}{M} \sum\limits_{j=1}^{M} V_{D}(x,U_{j})] & = \frac{1}{M} \sum\limits_{j=1}^{M} \mathbb{E}_{U_{j}} [V_{D}(x,U_{j})]\\
    & = \frac{1}{M}\sum\limits_{j=1}^{M} \mathbb{E}_{U}[V_{D}(x,U)]\\
    & = \mathbb{E}_{U}[V_{D}(x,U)]. 
\end{align*}
Then 
\begin{equation*}
    \frac{1}{M} \sum\limits_{j=1}^{M} V_{D}(x,U_{j}) - \mathbb{E}[V_{D}(x,U)] = \frac{1}{M}\sum\limits_{j=1}^{M} (V_{D}(x,U_{j}) - \mathbb{E}_{U}[V_{D}(x,U)]). 
\end{equation*}
By independence, for $j\neq k$, 
\begin{equation*}
    \mathbb{E}_{U_{j}, U_{k}} [(V_{D}(x,U_{j}) - \mathbb{E}_{U}[V_{D}(x,U)])(V_{D}(x,U_{k}) - \mathbb{E}_{U}[V_{D}(x,U)])] = 0. 
\end{equation*}
Therefore, 
\begin{align*}
    \mathbb{E}_{U_{1}, \dots, U_{M}}[(\frac{1}{M} \sum\limits_{j=1}^{M}V_{D}(x,U_{j}) - \mathbb{E}_{U}[V_{D}(x,U)])^{2}] &= \frac{1}{M^{2}} \sum\limits_{j=1}^{M} \sum\limits_{k=1}^{M} \mathbb{E}_{U_{1}, \dots, U_{M}}[(V_{D}(x,U_{j})\\& - \mathbb{E}_{U}[V_{D}(x,U)])(V_{D}(x,U_{k})-\mathbb{E}_{U}[V_{D}(x,U)])]\\
    & = \frac{1}{M^{2}} \sum\limits_{j=1}^{M} \mathbb{E}_{U_{j}} [(V_{D}(x,U_{j}) - \mathbb{E}_{U}[V_{D}(x,U)])^{2}]\\
    & = \frac{M}{M^{2}}\mathbb{E}_{U}[(V_{D}(x,U) - \mathbb{E}_{U}[V_{D}(x,U)])^{2}]\\
    & = \frac{1}{M}Var_{U}(V_{D}(x,U)). 
\end{align*}
Since $V_{D}(x,U) \in [0,1]$, 
\begin{equation*}
    -\frac{1}{2} \leq V_{D}(x,U) - \frac{1}{2} \leq \frac{1}{2}. 
\end{equation*}
So, 
\begin{equation*}
    (V_{D}(x,U)-\frac{1}{2})^{2} \leq \frac{1}{4}. 
\end{equation*}
Now, 
\begin{align*}
    Var_{U}(V_{D}(x,U)) & = \mathbb{E}_{U}[V_{D}(x,U)^{2}] - (\mathbb{E}_{U}[V_{D}(x,U)])^{2}\\
    & = \mathbb{E}_{U}[(V_{D}(x,U) - \frac{1}{2})^{2}] - (\mathbb{E}_{U}[V_{D}(x,U)] - \frac{1}{2})^{2}\\
    &\leq \mathbb{E}_{U}[(V_{D}(x,U) - \frac{1}{2})^{2}]\\
    & \leq \frac{1}{4}. 
\end{align*}
Then by the Cauchy-Schwarz inequality 
\begin{align*}
\mathbb{E}_{U_{1}, \dots, U_{M}}|\frac{1}{M}\sum\limits_{j=1}^{M}V_{D}(x,U_{j})-\mathbb{E}_{U}[V_{D}(x,U)]| & \leq \sqrt{\mathbb{E}_{U_{1},\dots,U_{M}} [(\frac{1}{M}\sum\limits_{j=1}^{M} V_{D}(x,U_{j})  - \mathbb{E}_{U}[V_{D}(x,U)])^{2}]}\\
&= \sqrt{\frac{1}{M} \operatorname{Var}_{U}(V_{D}(x,U))}\\
& \leq \frac{1}{2\sqrt{M}}.
\end{align*}
Conditioning on $D$, by the triangle inequality, 
\begin{equation*}
    |\frac{1}{M}\sum\limits_{j=1}^{M} V_{D}(x,U_{j}) - \eta_{0}(x)| \leq |\frac{1}{M} \sum\limits_{j=1}^{M} V_{D}(x,U_{j}) - \mathbb{E}_{U}[V_{D}(x,U)]| + |\mathbb{E}_{U}[V_{D}(x,U)] - \eta_{0}(x)|. 
\end{equation*}
Then taking expectations gives, 
\begin{align*}
    \mathbb{E}_{U_{1}, \dots, U_{M}}|\frac{1}{M}\sum\limits_{j=1}^{M} V_{D}(x,U_{j})-\eta_{0}(x)| &\leq \mathbb{E}_{U_{1}, \dots, U_{M}}|\frac{1}{M}\sum\limits_{j=1}^{M} V_{D}(x,U_{j}) - \mathbb{E}_{U}[V_{D}(x,U)]| \\&+ |\mathbb{E}_{U}[V_{D}(x,U)] - \eta_{0}(x)|\\
    &\leq \frac{1}{2\sqrt{M}} + |\mathbb{E}_{U}[V_{D}(x,U)] - \eta_{0}(x)|. 
\end{align*}
Then taking the expectation over $X$ yields, 
\begin{align*}
    \mathbb{E}_{X,U_{1}, \dots, U_{M}} [|\frac{1}{M}\sum\limits_{j=1}^{M} V_{D}(X,U_{j}) - \eta_{0}(X)||D] & \leq \frac{1}{2\sqrt{M}} +\mathbb{E}_{X}[|\mathbb{E}_{U}[V_{D}(X,U)] - \eta_{0}(X)||D]\\
    &\leq \frac{1}{2\sqrt{M}} + \int\limits_{\mathcal{X}} TV(P_{x}, Q_{D,x})\mu(dx)\\& + \int\limits_{\mathcal{X}} \int \limits_{\mathcal{Z}}|s_{D}(x,z) - \eta_{1}(x,z)|P_{x}(dz)\mu(dx). 
\end{align*}
Define 
\begin{equation*}
    g_{D,M}(x,u_{1}, \dots, u_{M}) = \boldsymbol{1}\{ \frac{1}{M} \sum\limits_{j=1}^{M} V_{D}(x,u_{j}) \geq \frac{1}{2} \}. 
\end{equation*}
Since $g_{D,M}$ is a measurable binary classifier only depending on $D, X$, and $(U_{1}, \dots, U_{M})$, Result 1 implies 
\begin{equation*}
    R_{D}(g_{D,M}) - R_{X}^{*} = \mathbb{E}_{X,U_{1}, \dots, U_{M}}[|2\eta_{0}(X)-1|\boldsymbol{1}\{ g_{D,M}(X, U_{1}, \dots, U_{M}) \neq g_{X}^{*}(X) \}|D]. 
\end{equation*}
Then, from the start of this proof, 
\begin{equation*}
    |2\eta_{0}(X)-1| \boldsymbol{1}\{ g_{D,M}(X, U_{1}, \dots, U_{M}) \neq g_{X}^{*}(X) \} \leq 2|\frac{1}{M}\sum\limits_{j=1}^{M}V_{D}(X,U_{j}) - \eta_{0}(X)|. 
\end{equation*}
Hence, 
\begin{align*}
    0 \leq R_{D}(g_{D,M}) - R_{X}^{*} &\leq 2[\mathbb{E}_{X,U_{1},  \dots, U_{M}}|\frac{1}{M}\sum\limits_{j=1}^{M} V_{D}(X,U_{j}) - \eta_{0}(X)||D]\\
    &\leq \frac{1}{\sqrt{M}} + 2\int\limits_{\mathcal{X}} TV(P_{x}, Q_{D,x})\mu(dx) +2\int\limits_{\mathcal{X}} \int \limits_{\mathcal{Z}}|s_{D}(x,z) - \eta_{1}(x,z)|P_{x}(dz)\mu(dx). 
\end{align*}
By result 1, 
\begin{equation*}
    R_{D}(g_{D,M}) - R_{XZ}^{*} = (R_{D}(g_{D,M})- R_{X}^{*}) + R_{X}^{*} - R_{XZ}^{*}. 
\end{equation*}
Finally, 
\begin{align*}
    R_{X}^{*} - R_{XZ}^{*} &\leq R_{D}(g_{D,M}) - R_{XZ}^{*} \\ &\leq R_{X}^{*} - R_{XZ}^{*} +  2\int\limits_{\mathcal{X}} TV(P_{x}, Q_{D,x})\mu(dx)\\& + 2\int\limits_{\mathcal{X}} \int \limits_{\mathcal{Z}}|s_{D}(x,z) - \eta_{1}(x,z)|P_{x}(dz)\mu(dx) + \frac{1}{\sqrt{M}}. 
\end{align*}
\subsection{Proof of Corollary 1}
Again consider some $V_{D}(X,U)$. Define 
\begin{align*}
    \ell(y,s) &:= -y\log(s)-(1-y) log(1-s),\\
    h(p) &:= -p\log(p) - (1-p)\log(1-p) \text{ with } 0\log(0):= 0,\\
    C_{D} &:= \mathbb{E}_{X,Y,U}[\ell(Y,V_{D}(X,U))|D], \text{ and } \\
    H_{X} &:= \mathbb{E}_{X}[h(\eta_{0}(X))]. 
\end{align*}
Then by the law of total expectation and $\mathbb{E}[Y|X,D,U] = \eta_{0}(X)$,
\begin{align*}
    C_{D}-H_{X} &= \mathbb{E}_{X,Y,U}[\ell (Y,V_{D}(X,U))|D] - \mathbb{E}_{X}[h(\eta_{0}(X))] \\
    &= \mathbb{E}_{X,U}[\mathbb{E}[\ell(Y, V_{D}(X,U))|X,D,U]|D]- \mathbb{E}_{X,U}[h(\eta_{0}(X))|D]\\
    &= \mathbb{E}_{X,U}[-\eta_{0}(X)\log(V_{D}(X,U)) - (1-\eta_{0}(X))\log(1-V_{D}(X,U))|D] - \mathbb{E}_{X,U}[h(\eta_{0}(X))|D]\\
    &= \mathbb{E}_{X,U}[-\eta_{0}(X)\log(V_{D}(X,U)) - (1-\eta_{0}(X))\log(1-V_{D}(X,U)) - h(\eta_{0}(X))|D]\\
    &= \mathbb{E}_{X,U}[-\eta_{0}(X) \log(V_{D}(X,U)) - (1-\eta_{0}(X))\log(1-V_{D}(X,U))\\& + \eta_{0}(X)\log(\eta_{0}(X)) + (1-\eta_{0}(X))\log(1-\eta_{0}(X))|D]\\
    & = \mathbb{E}_{X,U}[\eta_{0}(X) [\log(\eta_{0}(X)) - \log(V_{D}(X,U))]\\& + (1-\eta_{0}(X))[\log(1-\eta_{0}(X)) - \log(1-V_{D}(X,U))]|D]\\
    &= \mathbb{E}_{X,U} [KL(\eta_{0}(X)||V_{D}(X,U))|D], 
\end{align*}
where 
\begin{equation*}
    KL(p||q):=p\log(\frac{p}{q}) + (1-p)\log(\frac{1-p}{1-q}). 
\end{equation*}
By Pinsker's inequality, for any $p,q \in [0,1]$. 
\begin{equation*}
    KL(p|q) \geq 2TV(\operatorname{Bern}(p),\operatorname{Bern}(q))^2 = 2(p-q)^{2}.
\end{equation*}
Now consider the classifier defined as 
\begin{equation*}
    g_{D}^{V}(x,u) = \boldsymbol{1}\{ V_{D}(x,u) \geq \frac{1}{2} \}. 
\end{equation*}
We will prove that 
\begin{equation*}
    R_{D}(g_{D}^{V}) - R_{X}^{*} \leq \sqrt{2(C_{D}-H_{X})}. 
\end{equation*}
If $C_{D} = +\infty$, then the inequality is immediate. Otherwise, proceed as follows. By the inequality established above 
\begin{equation*}
    2(V_{D}(X,U) - \eta_{0}(X))^{2} \leq KL(\eta_{0}(X)||V_{D}(X,U)). 
\end{equation*}
Taking expectations gives 
\begin{equation*}
    2\mathbb{E}_{X,U} [(V_{D}(X,U) - \eta_{0}(X))^{2}|D] \leq \mathbb{E}_{X,U}[KL(\eta_{0}(X)||V_{D}(X,U))|D] = C_{D} - H_{X}. 
\end{equation*}
Then by the Cauchy-Schwarz inequality, 
\begin{align*}
    \mathbb{E}_{X,U}[|V_{D}(X,U) - \eta_{0}(X)||D] &= \mathbb{E}_{X,U}[|V_{D}(X,U)- \eta_{0}(X)\cdot 1|D]\\
    & \leq \sqrt{\mathbb{E}_{X,U}[(V_{D}(X,U) - \eta_{0}(X))^{2}|D]} \sqrt{\mathbb{E}_{X,U}[1^{2}|D]}\\
    &\leq \sqrt{\frac{C_{D}-H_{X}}{2}}. 
\end{align*}
Then by the proof of Result 1, 
\begin{align*}
    0  \leq R_{D}(g_{D}^{V}) - R_{X}^{*} &= \mathbb{E}_{X,U}[|2\eta_{0}(X) -1|\boldsymbol{1}\{ g_{D}^{V}(X,U) \neq g_{X}^{*}(X) \}|D]\\
    &\leq 2\mathbb{E}_{X,U}[|V_{D}(X,U) - \eta_{0}(X)||D], \text{ by the proof of Result 2,}\\
    & \leq 2\sqrt{\frac{C_{D}-H_{X}}{2}}\\
    &= \sqrt{2(C_{D}-H_{X})}. 
\end{align*}
Observe that 
\begin{equation*}
    \frac{\partial ^{2}}{\partial s^{2}} \ell (y,s) = \frac{y}{s^{2}} + \frac{1-y}{(1-s)^{2}} \geq 0. 
\end{equation*}
Then Jensen's inequality gives 
\begin{equation*}
    \ell(Y, \frac{1}{M}\sum\limits_{j=1}^{M}V_{D}(X,U_{j})) \leq \frac{1}{M} \sum\limits_{j=1}^{M} \ell(Y, V_{D}(X,U_{j})). 
\end{equation*}
Taking expectations gives 
\begin{align*}
    \mathbb{E}_{X,Y, U_{1}, \dots, U_{M}} [\ell(Y, \frac{1}{M}\sum\limits_{j=1}^{M} V_{D}(X,U_{j} ))|D] & \leq \frac{1}{M}\sum\limits_{j=1}^{M} \mathbb{E}_{X,Y, U_{j}} [\ell(Y, V_{D}(X,U_{j}))|D]\\
    & = \frac{1}{M} \sum\limits_{j=1}^{M} C_{D}\\
    &= C_{D}.
\end{align*}
Therefore, 
\begin{equation*}
    R_{D}(g_{D,M}) - R_{X}^{*}  \leq \sqrt{2(C_{D}-H_{X})}. 
\end{equation*}
\subsection{Proof of Result 3}
Let $L(\theta) = C(V_{\theta}) + \lambda F(T_{\theta})$, where $C(V_{\theta})$ is the classification loss and $F(T_{\theta})$ is the forecasting loss. To prove the result we make two assumptions. First, the training loss approximates population loss uniformly. In particular, 
\begin{equation*}
    \sup\limits_{\theta \in \Theta} |\hat{L}_{n}(\theta) - L(\theta)| \leq \epsilon_{n}
\end{equation*}
with high probability. Secondly, training approximately minimizes the empirical objective. That is, 
\begin{equation*}
    \hat{L}_{n}(\hat{\theta}) \leq \inf\limits_{\theta \in \Theta} \hat{L}_{n}(\theta) + \rho_{n}. 
\end{equation*}
Consider the event $\{ \sup\limits_{\theta \in \Theta} |\hat{L}_{n}(\theta) - L(\theta)| \leq \epsilon_{n} \}$ which has probability at least $1-\delta$ by assumption. Then for every $\theta \in \Theta$, 
\begin{align*}
    L(\hat{\theta}) & \leq \hat{L}_{n}(\hat{\theta}) + \epsilon_{n}\\
    &\leq \hat{L}_{n}(\theta) + \rho_{n} + \epsilon_{n}, \text{ by the training assumption},\\
    &\leq L(\theta) + \rho_{n} + 2\epsilon_{n}, \text{ by the uniform deviation assumption}. 
\end{align*}
Hence 
\begin{equation*}
    L(\hat{\theta}) \leq \inf\limits_{\theta \in \Theta} L(\theta) + \rho_{n} + 2\epsilon_{n}. 
\end{equation*}
Let $V_{\hat{\theta}}$ denote the probability function trained on data $D$. Then, from Corollary 1, we know that 
\begin{equation*}
    C(V_{\theta}) - H_{X} = \mathbb{E}_{X,U}[KL(\eta_{0}(X)||V_{\theta}(X,U))] \geq 0. 
\end{equation*}
Furthermore, by definition, 
\begin{equation*}
    F_{\min} := \inf\limits_{\theta^{'} \in \Theta} F(T_{\theta^{'}}) \leq F(T_{\theta}). 
\end{equation*}
Since $\lambda \geq 0$, 
\begin{equation*}
    \lambda[F(T_{\theta}) - F_{\min}] \geq 0. 
\end{equation*}
Thus, 
\begin{equation*}
    C(V_{\theta}) - H_{X} + \lambda[F(T_{\theta}) - F_{\min}] \geq 0 \text{ for every } \theta \in \Theta. 
\end{equation*}
Define
\begin{equation*}
    A_{\lambda} := \inf\limits_{\theta \in \Theta} \{ C(V_{\theta}) - H_{X} + \lambda [F(T_{\theta}) - F_{\min}] \} \geq 0. 
\end{equation*}
Then 
\begin{equation*}
    C(V_{\hat{\theta}}) + \lambda F(T_{\hat{\theta}}) \leq \inf\limits_{\theta \in  \Theta}\{ C(V_{\theta}) + \lambda F(T_{\theta}) \} + \rho_{n} + 2\epsilon_{n}. 
\end{equation*}
Subtracting $H_{X} + \lambda F_{\min}$ from both sides gives 
\begin{equation*}
    C(V_{\hat{\theta}}) - H_{X} + \lambda[F(T_{\hat{\theta}}) - F_{\min}]\leq A_{\lambda} +\rho_{n} +2\epsilon_{n}. 
\end{equation*}
Since $\lambda \geq 0$ and $F(T_{\hat{\theta}}) \geq F_{\min}$, 
\begin{equation*}
    C(V_{\hat{\theta}}) - H_{X} \leq A_{\lambda} + \rho_{n} + 2\epsilon_{n}. 
\end{equation*}
Then by corollary 1, 
\begin{align*}
    R_{D}(g_{D,M}) - R_{X}^{*} & \leq \sqrt{2(C(V_{D})-H_{X})}\\
    & = \sqrt{2(C(V_{\hat{\theta}}) - H_{X})}\\ 
    &\leq  \sqrt{2[A_{\lambda} + \rho_{n} + 2\epsilon_{n}]}. 
\end{align*}
Then 
\begin{align*}
    R_{D}(g_{D,M}) - R_{XZ}^{*} & = [R_{D}(g_{D,M}) - R_{X}^{*}] + [R_{X}^{*} - R_{XZ}^{*}]\\
    &\leq (R_{X}^{*}-R_{XZ}^{*}) + \sqrt{2(A_{\lambda} + \rho_{n} + 2\epsilon_{n})}
\end{align*}
Since the inequalities hold whenever $\sup\limits_{\theta \in \Theta} |\hat{L}_{n}(\theta) - L(\theta)| \leq \epsilon_{n}$ which has probability of at least $1-\delta$ over $D$, the bound also holds for every integer $M \geq 1$. 
\subsection{Proof of Corollary 2}
By assumption, $\ell_{\theta}(X_{1}, Z_{1}, Y_{1}), \dots, \ell_{\theta}(X_{n}, Z_{n}, Y_{n})$ are independent and identically distributed with $0 \leq \ell_{\theta}(X_{i}, Z_{i}, Y_{i}) \leq B$, for $i = 1, \dots, n$. Then by Hoeffding's inequality, for any $\epsilon>0$ and fixed $\theta \in \Theta$,
\begin{align*}
    \mathbb{P}(|\hat{L}_{n}(\theta) - L(\theta)| > \epsilon) &= \mathbb{P}(|\sum\limits_{i=1}^{n} (\ell_{\theta}(X_{i}, Z_{i}, Y_{i})-L(\theta))|> n\epsilon)\\
    &\leq 2\exp\{ - \frac{2(n\epsilon)^{2}}{nB^{2}} \}\\
    &= 2\exp \{ - \frac{2n\epsilon^{2}}{B^{2}}\}.
\end{align*}
Since $|\Theta| = K$, applying the union bound gives 
\begin{align*}
    \mathbb{P}(\sup\limits_{\theta \in \Theta}|\hat{L}_{n}(\theta) - L(\theta)|>\epsilon) &= \mathbb{P}(\bigcup\limits_{j=1}^{K} \{ |\hat{L}_{n}(\theta_{j}) - L(\theta_{j})| > \epsilon \})\\
    &\leq \sum\limits_{j=1}^{K} \mathbb{P}(|\hat{L}_{n}(\theta_{j}) - L(\theta_{j})|> \epsilon)\\
    &\leq \sum\limits_{j=1}^{K} 2\exp\{ - \frac{2n\epsilon^{2}}{B^{2}} \}\\
    &= 2K \exp \{ - \frac{2n\epsilon^{2}}{B^{2}} \}.
\end{align*}
Then setting 
\begin{equation*}
    \epsilon = \epsilon_{n} = B\sqrt{\frac{\log(\frac{2K}{\delta})}{2n}}
\end{equation*}
gives
\begin{equation*}
    \mathbb{P}(\sup\limits_{\theta \in \Theta} |\hat{L}_{n}(\theta) - L(\theta)| \leq \epsilon_{n}) \geq 1- \delta. 
\end{equation*}
\subsection{Proof of Proposition 1}
Define an estimator using the observed history-future pair: $\hat{\theta}= X_{1} - Z_{1}$. Then by substitution $\hat{\theta} = X_{1}- (X_{1}-\theta) = \theta$. So $\theta$ is exactly identified by one example. Then define the forecast and classification rule by 
\begin{equation*}
    \hat{Z}(x) := x- \hat{\theta} \text{ and } \hat{Y}(x):= \boldsymbol{1}\{ \hat{Z}(x) \geq 0 \}.
\end{equation*}
An independent test example is then generated by the same model: 
\begin{equation*}
    X \sim \operatorname{Unif}[0,1], Z:= X-\theta, \text{ and } Y:= \boldsymbol{1}\{ Z \geq 0 \}. 
\end{equation*}
Consequently 
\begin{equation*}
    \hat{Z}(X) = X- \hat{\theta} = X-\theta = Z \text{ and } \hat{Y}(X) = \boldsymbol{1}\{ Z \geq 0 \} = Y \qquad a.s.
\end{equation*}
That is, for every $\theta \in [0,1]$, 
\begin{equation*}
    \mathbb{P}_{\theta}(\hat{Y}(X) \neq Y)  = 0. 
\end{equation*}
Thus, the foreclasser achieves zero risk. It remains to establish a lower bound on the risk of the label-only model. Fix any label-only learner $A$. Assign the unknown parameter a uniform prior $\vartheta \sim \operatorname{Unif}[0,1]$. The future-supervised learner observes $D = ((X_{i}, Z_{i}, Y_{i}))_{i=1}^{n}$ whereas $A$ observes only $D_{n}^{\text{Lab}} = ((X_{i}, Y_{i}))_{i=1}^{n}$. By the law of total expectation, the expected risk is given by 
\begin{align*}
    \mathbb{E}_{\vartheta, D_{n}^{\text{Lab}}}[R_{\vartheta}(A(D_{n}^{\text{Lab}}))] &= \mathbb{E}_{\vartheta}[\mathbb{E}_{D_{n}^{\text{Lab}}|\vartheta}[R_{\vartheta}(A(D_{n}^{\text{Lab}}))]]\\
    &= \int\limits_{0}^{1} \mathbb{E}_{D_{n}^{\text{Lab}}|\vartheta = \theta}[R_{\theta}(A(D_{n}^{\text{Lab}}))] d\theta\\
    &= \int\limits_{0}^{1} \mathbb{E}_{D_{n}^{\text{Lab}}\sim (P_{\theta}^{XY})^{\otimes n}}[R_{\theta}(A(D_{n}^{\text{Lab}}))] d\theta. 
\end{align*}
Then define 
\begin{equation*}
    a_{n}:= \max (\{ X_{i}: 1 \leq i \leq n, Y_{i}=0 \} \cup \{ 0 \}), b_{n}:= \min(\{ X_{i}: 1 \leq i \leq n, Y_{i}=1 \} \cup \{1 \}), \text{ and } G_{n} = b_{n} - a_{n}. 
\end{equation*}
For the learner $A$, the label definition implies 
\begin{equation*}
    Y_{i} = 0 \implies \vartheta >X_{i} \text{ and } Y_{i} = 1 \implies \vartheta  \leq X_{i}. 
\end{equation*}
Therefore, each label-0 observation supplies a lower bound on $\theta$, and each label-1 observation supplies an upper bound. Combining these bounds implies $\theta$ must lie in 
\begin{equation*}
    \begin{cases}
        [0,b_{n}] \text{ if every }Y_{i} =1, \\
        (a_{n},b_{n}], \text{ otherwise.}
    \end{cases}
\end{equation*}
We also have that $G_{n}>0$ almost surely. To see this, $\vartheta \sim \operatorname{Unif}[0,1]$ so $0< \vartheta<1$ almost surely. Moreover, $\mathbb{P}(X_{i} = \vartheta)=0$ since the distribution of $X_{i}$ given $\vartheta = \theta$ is continuous. Therefore, from earlier in this proof, $X_{i} < \vartheta$ when $Y_{i}=0$ and $X_{i} > \vartheta$ when $Y_{i} =1$, almost surely. Then by the definition of $a_{n}$ and $b_{n}$, $a_{n} < \vartheta < b_{n}$ almost surely. Thus, $G_{n} = b_{n} - a_{n} >0$ almost surely. 
This implies 
\begin{align*}
    \mathbb{P}(Y_{1} = y_{1}, \dots, Y_{n} = y_{n} |X_{1}=x_{1}, \dots, X_{n} = x_{n}, \vartheta = \theta) &= \boldsymbol{1}\{ y_{i} = \boldsymbol{1}\{ x_{i} \geq \theta \} \text{ for every }i=1, \dots, n \}\\
    &= \prod\limits_{i=1}^{n}\boldsymbol{1}\{ y_{i} =1 \{ x_{i} \geq \theta \} \}\\
    &= (\prod\limits_{\substack{1 \leq i \leq n \\ y_i = 0}} \boldsymbol{1}\{ x_{i} < \theta \})(\prod\limits_{\substack{1 \leq i \leq n \\ y_i = 1}} \boldsymbol{1}\{ \theta \leq x_{i} \})\\
    &= \begin{cases}
        \boldsymbol{1}\{ 0 \leq \theta \leq b_{n} \}, \text{ if every }y_{i}=1\\
        \boldsymbol{1}\{ a_{n} < \theta \leq b_{n} \}, \text{ otherwise.}
    \end{cases}
\end{align*}
Then the posterior density is given by 
\begin{align*}
    f_{\vartheta|D_{n}^{\text{Lab}}}(\theta|D_{n}^{\text{Lab}}) &= \frac{f_{D_{n}^{\text{Lab}}|\vartheta}(D_{n}^{\text{Lab}}|\theta) f_{\vartheta}(\theta)}{\int\limits_{0}^{1}f_{D_{n}^{\text{Lab}}|\vartheta}(D_{n}^{\text{Lab}}|u) f_{\vartheta}(u)du}\\
    &= \frac{\mathbb{P}(Y_{1}=y_{1}, \dots, Y_{n} = y_{n}|X_{1} = x_{1}, \dots, X_{n} = x_{n}, \vartheta = \theta)}{\int\limits_{0}^{1}\mathbb{P}(Y_{1}=y_{1}, \dots, Y_{n} = y_{n}|X_{1} = x_{1}, \dots, X_{n} = x_{n}, \vartheta = u)du}\\
    &= \frac{\prod\limits_{i=1}^{n}\boldsymbol{1}\{ y_{i} = \boldsymbol{1}\{ x_{i} \geq \theta \}  \}}{\int\limits_{0}^{1}\prod\limits_{i=1}^{n} \boldsymbol{1}\{ y_{i} = \boldsymbol{1}\{x_{i} \geq u \} \}du}\\
    &= \begin{cases}
        \frac{\boldsymbol{1}\{ 0 \leq \theta \leq b_{n} \}}{\int\limits_{0}^{b_{n}}1 du}, \text{ if every }y_{i} =1\\
        \frac{\boldsymbol{1}\{ a_{n} < \theta \leq b_{n}\}}{\int\limits_{a_{n}}^{b_{n}}1 du}, \text{ otherwise,}
    \end{cases}\\
    &= \begin{cases}
        \frac{\boldsymbol{1}\{ 0 \leq \theta \leq b_{n} \}}{G_{n}} \text{ if every } y_{i} = 1,\\
        \frac{\boldsymbol{1}\{ a_{n} < \theta  \leq b_{n} \}}{G_{n}}, \text{ otherwise }. 
    \end{cases}
\end{align*}
Thus 
\begin{equation*}
    \vartheta|D_{n}^{\text{Lab}} \sim \operatorname{Unif}[a_{n}, b_{n}]. 
\end{equation*}
Observe that since $X|\vartheta=\theta \sim \operatorname{Unif}[0,1]$ for every $\theta$, the independent test history does not change the posterior distribution of $\vartheta$. This implies 
\begin{align*}
    \mathbb{P}(Y=1|X=x, D_{n}^{\text{Lab}}=D_{n}^{\text{Lab}}) &= \mathbb{P}(\vartheta \leq x|X=x, D_{n}^{\text{Lab}} = D_{n}^{\text{Lab}})\\
    &= \int\limits_{a_{n}}^{b_{n}} \boldsymbol{1}\{\theta \leq x \} \frac{1}{G_{n}}d \theta\\
    &= \begin{cases}
        0 , x \leq a_{n}\\
        \frac{1}{G_{n}} \int\limits_{a_{n}}^{x} \boldsymbol{1} d\theta, a_{n} < x < b_{n},\\
        \frac{1}{G_{n}}\int\limits_{a_{n}}^{b_{n}} 1 d\theta, x \geq b_{n},
    \end{cases}\\
    &= \begin{cases}
        0, x \leq a_{n}, \\
        \frac{x-a_{n}}{G_{n}}, a_{n}<x < b_{n}, \\
        1, x \geq b_{n}. 
    \end{cases}
\end{align*}
The Bayes rule chooses class 1 when $\mathbb{P}(Y=1|X=x, D_{n}^{\text{Lab}}=D_{n}^{\text{Lab}}) \geq 1/2$. That is, 
\begin{equation*}
    \frac{x-a_{n}}{G_{n}} \geq \frac{1}{2} \iff x \geq \frac{a_{n}+b_{n}}{2}.
\end{equation*}
Therefore the Bayes classifier is 
\begin{equation*}
    \boldsymbol{1}\{ x\geq \frac{a_{n}+b_{n}}{2} \}. 
\end{equation*}
From above, for $a_{n} < x< b_{n}$, the conditional error probabilities are $(x-a_{n})/G_{n}$ if we predict class 0 and $(b_{n}-x)/G_{n}$ if we predict class 1. Therefore, any decision has error at least the minimum of these two quantities. Then 
\begin{align*}
    \mathbb{E}[R_{\vartheta}(A(D_{n}^{\text{Lab}})) |D_{n}^{\text{Lab}} = D_{n}^{\text{Lab}}] &\geq \int\limits_{a_{n}}^{b_{n}} \min\{ \frac{x-a_{n}}{G_{n}}, \frac{b_{n}-x}{G_{n}} \}dx\\
    &= \int\limits_{a_{n}}^{\frac{a_{n}+b_{n}}{2}} \frac{x-a_{n}}{G_{n}} dx + \int\limits_{\frac{a_{n} + b_{n}}{2}}^{b_{n}} \frac{b_{n}-x}{G_{n}}dx\\
    &= \frac{1}{G_{n}}[\frac{(x-a_{n})^{2}}{2}]_{a_{n}}^{\frac{a_{n}+b_{n}}{2}} + \frac{1}{G_{n}}[-\frac{(b_{n}-x)^{2}}{2}]_{\frac{a_{n}+b_{n}}{2}}^{b_{n}}\\
    &= \frac{1}{2G_{n}}[(\frac{a_{n}+b_{n}}{2}-a_{n})^{2} + (b_{n} - \frac{a_{n}+b_{n}}{2})^{2}]\\
    &= \frac{1}{2G_{n}}[(\frac{b_{n}-a_{n}}{2})^{2} + (\frac{b_{n}-a_{n}}{2})^{2}]\\
    &= \frac{1}{2G_{n}}[\frac{G_{n}^{2}}{4} + \frac{G_{n}^{2}}{4}]\\
    &= \frac{G_{n}}{4}. 
\end{align*}
Observe that 
\begin{equation*}
    G_{n} = b_{n} - a_{n} = \int\limits_{0}^{1} \boldsymbol{1}\{ a_{n} < x < b_{n} \}dx.
\end{equation*}
Furthermore, consider fixed $x, \theta \in (0,1)$ and condition on $\vartheta = \theta$. If $x<\theta$ then $b_{n}> x$. This implies $a_{n} < x < b_{n}$ holds exactly when no label-0 input lies between $x$ and $\theta$. Similarly, if $x > \theta$ then $a_{n} <x$. This implies $a_{n} < x < b_{n}$ holds exactly when no label-1 input lies between $\theta$ and $x$. Finally, if $x=\theta$ and $\theta \in (0,1)$, then $a_{n}< x < b_{n}$ holds almost surely. 
Now
\begin{align*}
    \mathbb{P}(a_{n} < x < b_{n}|\vartheta = \theta) &= \mathbb{P}(X_{i} \notin (\min\{ x,\theta\}, \max\{ x , \theta \}) \text{ for every }i |\vartheta = \theta)\\
    &= \prod\limits_{i=1}^{n} \mathbb{P}(X_{i} \notin (\min\{ x,\theta \}, \max\{ x, \theta \})|\vartheta = \theta)\\
    &= \prod\limits_{i=1}^{n}[1-(\max\{ x, \theta \}-\min\{x,\theta \})]\\
    &= \prod\limits_{i=1}^{n} (1-|x-\theta|)\\
    &= (1-|x-\theta|)^{n}.
\end{align*}
Then by Tonelli's theorem 
\begin{align*}
    \mathbb{E}[G_{n}] &= \mathbb{E}[\int\limits_{0}^{1} \boldsymbol{1}\{ a_{n}<x<b_{n} \} dx]\\
    &=\int\limits_{0}^{1}\mathbb{E}[\boldsymbol{1}\{ a_{n}<x< b_{n} \}]dx\\
    &=\int\limits_{0}^{1} \mathbb{P}(a_{n} < x < b_{n})dx\\
    &= \int\limits_{0}^{1}\int\limits_{0}^{1}\mathbb{P}(a_{n} < x < b_{n}|\vartheta = \theta)d\theta dx\\
    &= \int\limits_{0}^{1} \int\limits_{0}^{1}(1-|x-\theta|)^{n}d\theta dx\\
    &= \int\limits_{0}^{1}\int\limits_{0}^{1} (1-|x-\theta|)^{n}dxd\theta \\
    &= \int\limits_{0}^{1}[\int\limits_{0}^{\theta}(1-\theta +x)^{n}dx + \int\limits_{\theta}^{1}(1-x+\theta)^{n}dx]d\theta\\
    &= \int\limits_{0}^{1}[[\frac{(1-\theta + x)^{n+1}}{n+1}]_{0}^{\theta} + [-\frac{(1-x+\theta)^{n+1}}{n+1}]_{\theta}^{1}]d\theta\\
    &= \int\limits_{0}^{1}[\frac{1-(1-\theta)^{n+1}}{n+1} + \frac{1-\theta^{n+1}}{n+1}] d\theta\\
    &= \frac{1}{n+1}\int\limits_{0}^{1}[2-(1-\theta)^{n+1}-\theta^{n+1}]d\theta\\
    &= \frac{1}{n+1}[2\theta + \frac{(1-\theta)^{n+2}}{n+2} - \frac{\theta^{n+2}}{n+2}]_{0}^{1}\\
    &= \frac{1}{n+1}[(2-\frac{1}{n+2})-\frac{1}{n+2}]\\
    &= \frac{1}{n+1}(2-\frac{2}{n+2})\\
    &= \frac{1}{n+1}\frac{2(n+1)}{n+2}\\
    &= \frac{2}{n+2}. 
\end{align*}
Then 
\begin{align*}
    \int\limits_{0}^{1} \mathbb{E}_{D_{n}^{\text{Lab}} \sim (P_{\theta}^{XY})^{\otimes n}}[R_{\theta}(A(D_{n}^{\text{Lab}}))]d \theta &= \mathbb{E}_{\vartheta, D_{n}^{\text{Lab}}}[R_{\vartheta}(A(D_{n}^{\text{Lab}}))]\\
    &= \mathbb{E}_{D_{n}^{\text{Lab}}}[\mathbb{E}[R_{\vartheta}(A(D_{n}^{\text{Lab}}))|D_{n}^{\text{Lab}}]]\\
    &\geq \mathbb{E}_{D_{n}^{\text{Lab}}}[\frac{G_{n}}{4}]\\
    &= \frac{1}{4}\mathbb{E}[G_{n}]\\
    &= \frac{1}{4}\cdot \frac{2}{n+2}\\
    &= \frac{1}{2(n+2)}. 
\end{align*}
\end{document}